\documentclass{article}

\usepackage{microtype}
\usepackage{graphicx}
\usepackage{subcaption}
\usepackage{booktabs} 

\usepackage{hyperref}

\usepackage[preprint]{icml2026}

\usepackage{amsmath}
\usepackage{amssymb}
\usepackage{mathtools}
\usepackage{amsthm}
\usepackage{algorithm}
\usepackage{algorithmic}
\usepackage{amsfonts}
\usepackage{bm}
\usepackage{stfloats}
\usepackage{tcolorbox}
\usepackage{graphicx}
\usepackage{etoc}

\usepackage[capitalize,noabbrev]{cleveref}

\theoremstyle{plain}
\newtheorem{theorem}{Theorem}[section]

\theoremstyle{definition}
\newtheorem{definition}[theorem]{Definition}

\theoremstyle{remark}

\usepackage[textsize=tiny]{todonotes}

\usepackage{xspace, graphicx, amssymb, caption, multirow, overpic, textpos, wrapfig, pifont, array, xcolor, subcaption}
\usepackage{enumitem}
\usepackage[table]{xcolor}
\usepackage{epigraph}
\usepackage{hyperref}

\icmltitlerunning{Reliable Thinking with Images}

\definecolor{hr}{gray}{0.7}  
\definecolor{dt}{HTML}{ADCAD8}  

\newlength\savewidth
\newcommand{\tablestyle}[2]{\setlength{\tabcolsep}{#1}\renewcommand{\arraystretch}{#2}\centering\footnotesize}
\renewcommand{\paragraph}[1]{\vspace{1.25mm}\noindent\textbf{#1}}

\begin{document}

\twocolumn[
  \icmltitle{Reliable Thinking with Images}


    \icmlsetsymbol{corres}{$\dagger$}

  \begin{icmlauthorlist}
    \icmlauthor{Haobin Li}{scu}
    \icmlauthor{Yutong Yang}{scu}
    \icmlauthor{Yijie Lin}{scu}
    \icmlauthor{Xiang Dai}{sci}
    \icmlauthor{Mouxing Yang}{corres,scu}
    \icmlauthor{Xi Peng}{corres,scu,nkl}
  \end{icmlauthorlist}
\begin{center}
\url{https://github.com/XLearning-SCU/Reliable_TWI}
\end{center}
  \icmlaffiliation{scu}{College of Computer Science, Sichuan University}
  \icmlaffiliation{sci}{Southwest China Institute of Electronic Technology}
  \icmlaffiliation{nkl}{National Key Laboratory of Fundamental Algorithms and Models for Engineering Simulation, Sichuan University}
  \icmlcorrespondingauthor{Mouxing Yang, Xi Peng}{{yangmouxing, pengx.gm}@gmail.com}

  \vskip 0.3in
]



\printAffiliationsAndNotice{}  

\begin{abstract}
As a multimodal extension of Chain-of-Thought (CoT), Thinking with Images (TWI) has recently emerged as a promising avenue to enhance the reasoning capability of Multi-modal Large Language Models (MLLMs), which generates interleaved CoT by incorporating visual cues into the textual reasoning process. However, the success of existing TWI methods heavily relies on the assumption that interleaved image-text CoTs are faultless, which is easily violated in real-world scenarios due to the complexity of multimodal understanding. In this paper, we reveal and study a highly-practical yet under-explored problem in TWI, termed Noisy Thinking (NT). Specifically, NT refers to the imperfect visual cues mining and answer reasoning process. As the saying goes, ``One mistake leads to another'', erroneous interleaved CoT would cause error accumulation, thus significantly degrading the performance of MLLMs. To solve the NT problem, we propose a novel method dubbed Reliable Thinking with Images (RTWI). In brief, RTWI estimates the reliability of visual cues and textual CoT in a unified text-centric manner and accordingly employs robust filtering and voting modules to prevent NT from contaminating the final answer. Extensive experiments on seven benchmarks verify the effectiveness of RTWI against NT. 
\end{abstract}

\section{Introduction}

\begin{figure*}[t]
    \centering
    \includegraphics[width=0.93\linewidth]{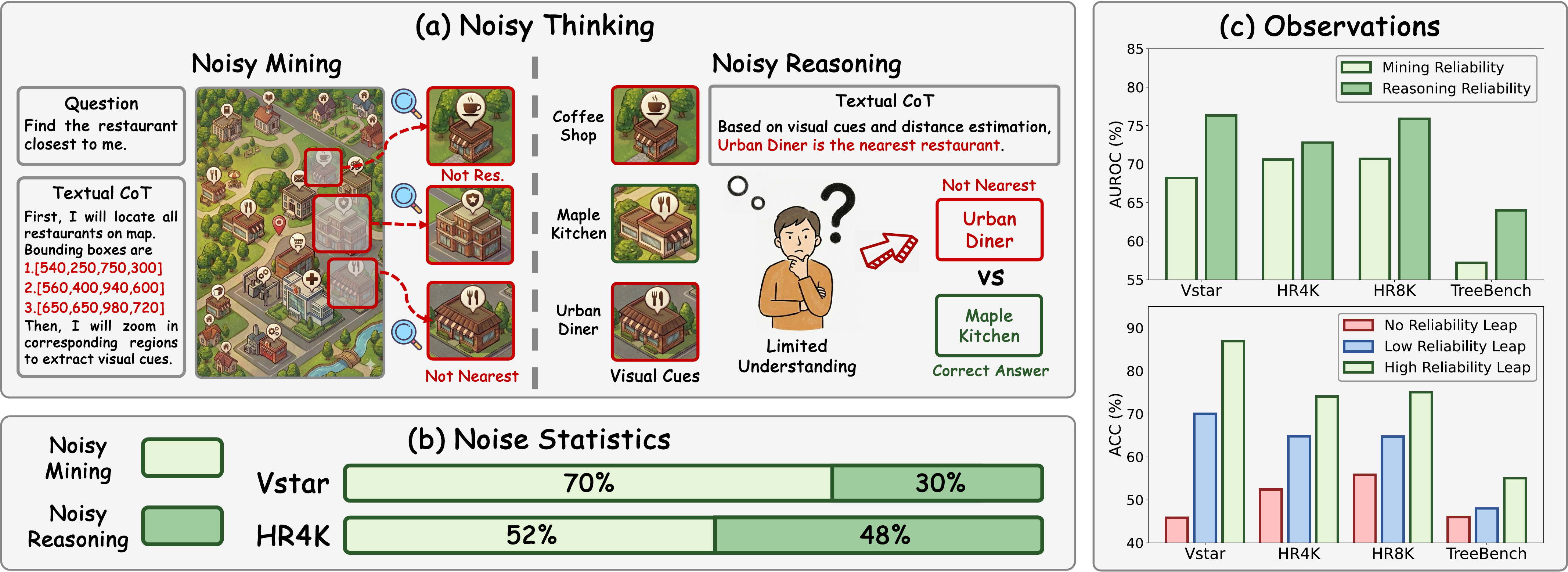}
    \caption{(a) \textbf{Noisy Thinking:} the TWI paradigm would inevitably suffer from the noisy thinking problem at both the mining and reasoning stages. On the one hand, MLLMs would extract task-agnostic or inaccurate visual cues once the textual CoT generates unreliable tool-calling instructions. 
    Clearly, it is difficult to derive the right answer with incorrect visual cues, \textit{e.g.}, the question is ``Find the restaurant closest to me.'', but the visual cues either do not depict restaurants or correspond to the non-closest one.
    On the other hand, even desirable information exists in acquired visual cues, MLLM might generate erroneous textual CoT and then the incorrect answer, \textit{e.g.}, failing to identify the nearest restaurant due to limited distance estimation.
    (b) \textbf{Noise Statistics:} we investigate the TWI cases in the Vstar and HR4k datasets and observe that NT in either mining or reasoning would lead to wrong answers in real-world scenarios.
    (c) \textbf{Observations:} 
    we estimate the reliability of the two stages in the TWI process using tool-invocation and reasoning tokens, and then adopt the AUROC metric to quantify the relationship between reliability and reasoning accuracy.
    As a result, one could observe that traces with correct answers tend to exhibit higher dual-stage reliabilities and reliability leap from mining stage to reasoning stage.
    }
    \label{fig: fig1}
\end{figure*}

\renewcommand{\epigraphflush}{flushleft} 
\setlength{\epigraphwidth}{1\columnwidth} 
{
\epigraph{\textit{The devil is in the details.}}{\textit{--- Ludwig Mies van der Rohe}}
}\vspace{-3mm}
Chain-of-Thought (CoT)~\cite{cot1,cot2} has become the dominant paradigm for reasoning with Large Language Model (LLM)~\cite{llm1,llm2}, which solves complex problems by generating a sequence of textual reasoning steps.
Inspired by this, recent advances in Multimodal Large Language Model (MLLM)~\cite{mllm1,mllm2} primarily adopt the ``reasoning within language'' paradigm over textual and visual inputs to enhance multimodal reasoning capability.
However, such a paradigm simply encodes the whole visual input as the static context, making it difficult to exploit task-relevant visual information.
In other words, visual inputs are always rich and redundant, which might obscure the critical details necessary for accurate answer derivation.
As a remedy, endowing MLLM with the capability to ``thinking with images''~\cite{openaio3,deepeyes,minio3} has recently emerged as a promising paradigm, which actively manipulates visual information in intermediate steps and thus forms interleaved image-text CoTs.
Specifically, Thinking With Images (TWI) paradigm typically consists of two stages: 
i) \textbf{Cue Mining}: MLLMs inspect the whole visual inputs and generate textual CoT to call external tools for image manipulation, thereby acquiring task-relevant visual cues;
ii) \textbf{Answer Reasoning}: with the desirable visual cues, MLLMs have incentives to derive the correct answer and solve the task.

Despite the promising performance of the existing TWI approaches, their success heavily relies on the assumption that both the cue mining and answer reasoning stages are faultless.
However, in the real-world TWI applications, such an ideal assumption is daunting and even impossible to satisfy, leading to the \textbf{Noisy Thinking (NT)} problem.
Specifically, as shown in Fig.~\ref{fig: fig1}(a), NT refers to the noisy cue mining and noisy answer reasoning in the TWI process.
Specifically,
i) \textbf{Noisy Mining}: akin to finding needles in a haystack, it is inevitable to extract task-irrelevant or coarse-grained cues from redundant visual inputs;
ii) \textbf{Noisy Reasoning}: even with desirable visual cues, the inherent limitations of MLLMs in multimodal understanding would contaminate the textual CoT and lead to incorrect answer.
As the saying goes, ``One mistake leads to another'', either the undesirable visual cues in the mining stage or the erroneous textual CoT in the reasoning stage would propagate along the TWI process, thus significantly degrading the performance.

To handle the NT problem, the most promising solution might be the Test-Time Scaling (TTS)~\cite{tts1,tts2}, which improves reasoning accuracy by sampling multiple reasoning paths and deriving answers via search-~\cite{search-tts1,search-tts2} or consistency-based~\cite{sc,self-certainty} mechanisms.
However, most existing TTS methods are specifically designed for ``reasoning with language'' paradigm in LLM, while overlooking the complexity of ``thinking with images'' paradigm in MLLM.
Specifically, they are intractable for addressing the NT problem due to the following two reasons.
On the one hand, unlike discrete and tokenized text, visual signals are continuous and unsegmented, making visual uncertainty estimation substantially more  challenging than token-level uncertainty modeling. 
Moreover, visual uncertainty is not necessarily aligned with cue uncertainty, as it might indicate image quality yet fail to reveal whether the mined cue is task-relevant or not.
On the other hand, it remains under-explored how inaccurate visual cues propagate into and distort subsequent textual reasoning, not to mention how to achieve NT-robust TTS. 

Instead of exhaustingly modeling visual uncertainty, we argue that the image-manipulation textual CoT itself could implicitly serve as a natural proxy for visual cue reliability, thus enabling a dual-stage robust TTS framework to tackle NT in the TWI process.
To support our claims, we present two observations in Fig.~\ref{fig: fig1}(c).
To be specific,
i) Reliability Correlation: both the reliabilities of the visual
cues in the mining stage and the textual CoT in the reasoning stage are correlated with final performance, \textit{i.e.}, reliable visual cues and answer reasoning steps would facilitate the obtaining of correct answers;
ii) Reliability Leap: the correct visual cues tend to induce more reliability leap from the mining to the reasoning stage.
Intuitively, akin to human cognition, the acquisition of correct visual cues would boost the confidence of the subsequent reasoning process.

Based on the above discussions and observations, we propose a TTS-based TWI method to achieve robustness against NT, dubbed Reliable Thinking With Images (RTWI), which samples multiple thinking traces for mining and deriving trustworthy answer.
In brief, RTWI adopts a text-centric reliability estimation mechanism, which could identify noisy mining and reasoning in a unified manner.
With formulated reliability, RTWI employs a dual-stage filtering module to discard unreliable thinking traces with self-adaptive thresholds, thereby preventing NT from contaminating answer derivation.
Finally, RTWI conducts NT-robust voting over the remaining traces by resorting to the reliability leap characteristic, which contributes to deriving trustworthy answer. 
In summary, the major contributions and novelties of this work are given as follows.
\begin{itemize}
    \item We reveal and study a novel and practical problem in TWI, termed Noisy Thinking (NT). In brief, NT refers to the noisy mining and reasoning stages in the TWI process, which would propagate errors through the subsequent reasoning and significantly degrade MLLM performance.
    \item To achieve robust TWI against NT, we propose a novel method termed RTWI. In brief, RTWI estimates the reliability of visual cues and textual CoT in a unified text-centric manner and then mitigates the negative impact of NT through dual-stage filtering and voting.
    \item The proposed RTWI could not only mitigate NT on offline-generated traces, but also enable reliable early stopping during online generation and thus significantly improve efficiency.
    Extensive experiments on seven benchmarks under offline and online settings demonstrate the effectiveness of RTWI against noisy thinking compared with state-of-the-art methods.
\end{itemize}

\section{Related Work}
In this section, we provide a brief review of two topics highly related to this work, including thinking with images and test-time scaling.

\subsection{Thinking with Images}
Thinking with images aims to solve complex multimodal questions by actively manipulating images (\textit{e.g.}, zooming into regions of interest) to support subsequent reasoning, which is popularized by OpenAI o3~\cite{openaio3}.
From the perspective of how to endow MLLMs with the TWI capability, existing approaches could be categorized into two groups:
i) prompt-based methods~\cite{dyfo,zoomeye}, which carefully design prompts to coordinate external tools for extracting visual cues without updating model parameters;
ii) training-based methods~\cite{thyme,deepeyes,minio3}, which conduct supervised fine-tuning on predefined trajectories to learn structured reasoning patterns, or adopt reinforcement learning with outcome-based rewards to explore visual manipulation and reasoning policies autonomously.

Among existing approaches, DRIM~\cite{DRIM} is the most relevant to our work.
In brief, DRIM leverages a carefully-designed reinforcement learning scheme to strengthen the self-reflection capacity of MLLMs during the TWI process.
In contrast, rather than modifying training to improve robustness, we propose a reliable TTS mechanism at inference time, which is orthogonal to DRIM and could serve as a plug-and-play solution for a wide spectrum of TWI models, enabling more robust thinking.

\subsection{Test-time Scaling}
Test-time Scaling has become the dominant paradigm to improve reasoning accuracy in LLM, revealing a scaling law that more compute leads to better performance, as supported by representative models include OpenAI o1~\cite{openaio1}, DeepSeek R1~\cite{deepseekr1}, Grok-4~\cite{grok4}, and Qwen3~\cite{qwen3}.
Based on the ways to utilize extra test-time compute, the existing TTS approaches could be divided into three categories:
i) self-consistency methods~\cite{sc}, which sample multiple reasoning trajectories and aggregate them via majority voting;
ii) early stopping methods~\cite{esc,asc}, which seek better accuracy-compute trade-offs by adaptively terminating further token generation once the existing traces reach consensus;
iii) robust reasoning methods~\cite{deepconf,self-certainty}, which dynamically filter out low-confidence traces or scale low-confidence reasoning steps to improve reasoning accuracy.

The major differences between existing TTS methods and this work are given below.
On the one hand, existing TTS methods are carefully designed for textual CoT in ``reasoning with language'' paradigm, whereas our approach addresses the more challenging interleaved CoT in ``thinking with images'' paradigm.
On the other hand, to the best of our knowledge, no prior work has explored how to identify undesirable visual cues, not to mention mitigating the negative impacts of NT in the TWI process.

\begin{figure*}[t]
    \centering
    \includegraphics[width=0.95\linewidth]{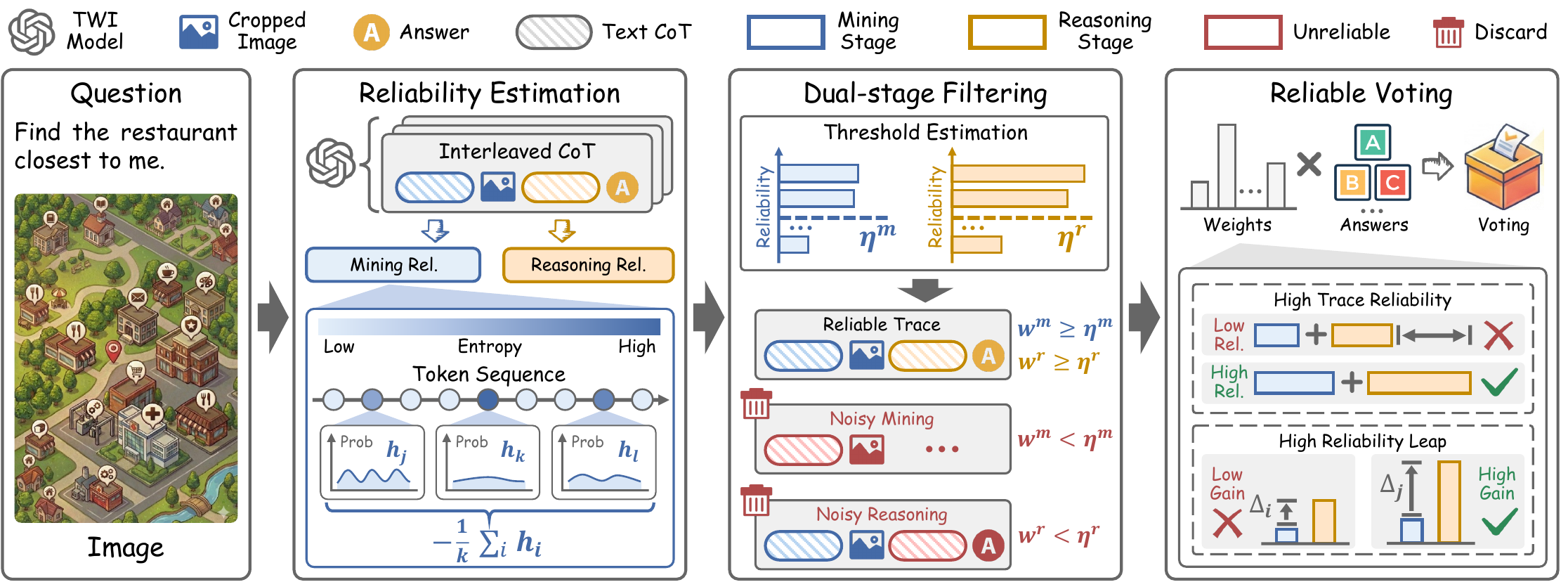}
    \caption{Overview of our method RTWI. For clarity, we take single-turn cue mining as a showcase and denote stage reliability $w(t^s)$ as $w^s$ with $s\in \{m,r\}$.
    Given a multimodal question, RTWI first generates multiple interleaved traces and estimates the reliability of visual cues in the mining stage and textual CoT in the reasoning stage. After that, RTWI identifies and filters the unreliable traces with self-adaptive thresholds. 
    Finally, RTWI assigns higher weights to trustworthy traces based on two carefully-designed principles and then aggregates them to derive the answer.}
    \label{fig: fig2}
\end{figure*}

\section{Method}
In this section, we introduce Reliable Thinking With Images (RTWI) to tackle NT for TWI.
In Section~\ref{sec: problem_formulation}, we present the formal definition of the TWI process.
In Section~\ref{sec: reliability_estimation}, we propose a text-centric reliability estimation mechanism to identify NT. 
In Section~\ref{sec: dual_stage_filtering}-\ref{sec: reliable_voting}, we introduce dual-stage filtering and reliable voting modules to achieve robustness against NT.

\subsection{Problem Formulation}
\label{sec: problem_formulation}
Without loss of generality, we take the one thinking trace $t$ as a showcase to elaborate on the TWI process.
Let $\mathcal{V}$ denotes the space of all possible textual outputs (\textit{e.g.}, the vocabulary of text tokens), $\mathcal{I}$ denotes the space of all possible intermediate visual cues, and $\Theta$ represents the parameters of the MLLM.
Given a question $Q$ with an initial image $I$, TWI aims to derive the answer by generating a sequence of interleaved image-text CoT $t^{m}=\big\{(\mathbf{x}_i^{\text{txt}}, \mathbf{x}_i^{\text{vis}})\big\}_{i=1}^{N-1}$ in the mining stage and subsequently facilitating the textual CoT $t^{r}=\{x_{N}^{\text{txt}}\}$ in the reasoning stage, where $N$ denotes the total number of turns, $\mathbf{x}_i^{\text{txt}}\in \mathcal{V}$ and $\mathbf{x}_i^{\text{vis}}\in \mathcal{I}$ denote the textual CoT and the corresponding visual cues at $i$-th turn, respectively. 
Specifically,

\textbf{Mining Stage ($i<N$):} the interleaved image-text CoT could be derived as
\begin{equation}
\begin{gathered}
    \mathbf{x}_i^{\text{txt}} \sim P(\,\cdot \mid S_i, I, Q; \Theta),\\
    \mathbf{x}_i^{\text{vis}}=f_{\text{tool}}(\mathbf{x}_i^{\text{txt}},I),
\end{gathered}
\label{eq: mining_stage}
\end{equation}
where $S_i=\big\{(\mathbf{x}_j^{\text{txt}}, \mathbf{x}_j^{\text{vis}})\big\}_{j=1}^{i-1}$ denotes thinking history at $i$-th turn, $f_{\text{tool}}$ denotes the tool invocation (\textit{e.g.}, zooming in), $P(\,\cdot \mid \cdot \, ; \, \Theta)$ denotes the conditional distributions parameterized by MLLM, and $\sim$ indicates sampling process. 

\textbf{Reasoning Stage ($i=N$):} based on the extracted visual cues, the MLLM generates textual CoT to conduct comprehensive reasoning and derive the answer,
\begin{equation}
\begin{gathered}
    \mathbf{x}_N^{\text{txt}} \sim P(\,\cdot \mid S_N, I, Q; \Theta),\\
    A_t=f_{\text{ans}}(\mathbf{x}_N^{\text{txt}}),
\end{gathered}
\label{eq: reasoning_stage}
\end{equation}
where $A_t$ indicates the answer of $t$-th trace, $f_{\text{ans}}(\cdot)$ denotes the deterministic function that extracts the answer from the textual outputs.
As discussed in Introduction, existing TWI methods assume that both the mining and reasoning stages are faultless, which is often violated in real-world scenarios and thus leads to the NT problem.

To tackle the NT problem, we propose Reliable Thinking With Images (RTWI) as shown in Fig.~\ref{fig: fig2}, which consists of the reliability estimation module, the dually robust filtering module, and the reliable fusion module.
In the following, we will elaborate on each of them.

\begin{figure}[t]
    \centering
    \includegraphics[width=0.9\linewidth]{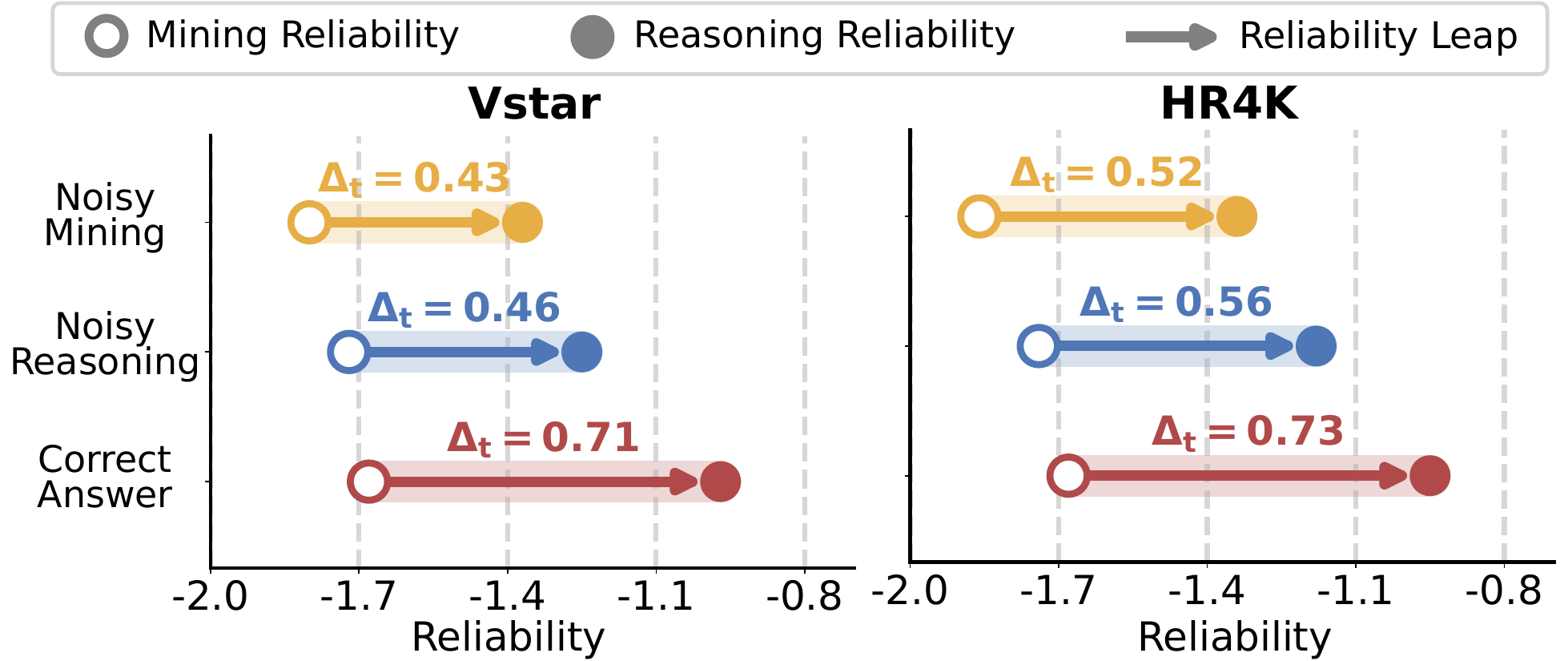}
    \caption{Failure case analyses. ``Noisy Mining'' and ``Noisy Reasoning'' indicate the underlying causes of incorrect answers.}
    \label{fig: observation_3}
\end{figure}

\subsection{Reliability Estimation}
\label{sec: reliability_estimation}
Recent studies~\cite{semantic_entropy,cisc} have demonstrated that the quality of thinking trace could be estimated from model-intrinsic signals without requiring external supervision, \textit{e.g.}, token confidence and trace distribution.
However, existing methods are carefully designed for ``reasoning with language'' paradigm, leaving an urgent need to identify both noisy visual cues and textual CoT as mentioned in Introduction.

To remedy this, instead of exhaustingly estimating visual uncertainty, we delve into the cue mining stage and reveal that the acquisition of visual cues is inherently reflected by the textual CoT.
In other words, reliable textual CoT for tool invocation could act as the precondition for desirable visual cues derivation.
Inspired by this, we propose a unified text-centric paradigm to estimate the reliability of both visual cues and textual CoT, which facilitates the identification of NT.
Specifically, we give a formal definition of stage reliability.
\begin{definition}[Stage Reliability]
    For a given trace $t$, the reliability of stage $t^{s} (s\in \{m,r\})$ is defined as
    \begin{equation}
        w(t^{s})=-\frac{1}{k}\sum_{i \in \mathcal{K}(\mathcal{H}\left(t^{s})\right)}h^s_i,
        \label{eq: stage_reliability}
    \end{equation}
    where $\mathcal{H}(t^{s})$ denotes the set of token entropies in $t^{s}$, $\mathcal{K}(\cdot)$ returns the indices of the Top-$k$ largest values, $w(t^{s})\leq0$ and $|w(t^{s})|$ is negatively-correlated with the stage reliability. 
\end{definition}
More specifically, the entropy of $i$-th token in $\mathcal{H}(t^{s})$ is defined as $h_{i}^s=-\sum_{j\in \mathcal{V}} p^s_{ij}\log p^s_{ij}$, where $\{p^s_{ij}\}_{j\in \mathcal{V}}$ denotes the predicted probability distribution produced by MLLM.
Such a reliability formulation embraces the following merits:
i) it avoids considering unimportant or redundant tokens by concentrating on pivotal decision points with higher entropies;
ii) the Top-$k$ selection yields more stable reliability estimation across mining and reasoning stages, regardless of CoT length (see more details in Appendix~\ref{appendix: trace_length}).
Intuitively, the unreliable textual CoT tend to exhibit low reliability, which could serve as a promising principle to identify NT.

To support our claims, we analyze the stage-wise reliabilities of the traces with either wrong or correct answers.
From the results in Fig.~\ref{fig: observation_3}, one could observe that errors in the current stage would propagate to the subsequent reasoning process, leading to cascading degradation of reliability.
In contrast, traces with correct answers exhibit high reliability across both stages, further verifying that reliability could serve as a principle for identifying NT.

\subsection{Dual-stage Filtering}
\label{sec: dual_stage_filtering}
To prevent unreliable traces from contaminating the answer, we propose to generate multiple thinking traces and prioritize trustworthy traces among them.
Specifically, the reliable thinking traces could be determined as follows,
\begin{equation}
    T_{\text{rel}}=\big\{ t \;\big|\; w(t^{m})\ge \eta^{m} \;\land\; w(t^{r})\ge \eta^{r}\big\},
    \label{eq: dual-stage filtering}
\end{equation}
where $T_{\text{rel}}$ denotes the set of reliable traces, $\eta^{m}$ and $\eta^{r}$ indicate the filtering thresholds in mining and reasoning stages, respectively.
Specifically, for each multimodal question, we sample warmup traces to derive stage-wise thresholds, \textit{i.e.},
\begin{equation}
\eta^{s}=\operatorname{Percentile}_{\alpha}
\big(\{\, w(t^{s}) \mid t \in T_{\text{sel}} \,\}\big),
\label{eq: dual-level thresholds}
\end{equation}
where $s\in \{m,r\}$, $T_{\text{sel}}$ represents selected traces for threshold estimation, $\operatorname{Percentile}_{\alpha}(\cdot)$ indicates that traces with the lowest $\alpha$ fraction of reliabilities are filtered at this stage.
Such behavior could exclude unreliable traces with potentially erroneous mining or reasoning, thus facilitating the derivation of the correct answer.
It is worth noting that the dual-stage filtering module could be easily integrated into existing TWI models under both online and offline settings, see more details in Alg.~\ref{alg: online}-\ref{alg: offline}.
More specifically, in the online pipeline, traces are generated one by one, and $T_{\text{sel}}$ denotes the subset of warmup traces used to estimate the stopping thresholds, which enable dynamic early stopping for subsequent unreliable traces within the budget.
In contrast, in the offline pipeline, all traces are pre-generated and $T_{\text{sel}}$ refers to the full set of generated traces used to estimate thresholds for post-hoc filtering.

\begin{table*}[t] 
\centering
\caption{Results on real-world high-resolution benchmarks under online setting. ``ACC'' and ``TSR'' denote the final accuracy and token saving ratio, respectively. The best results are marked in \textbf{bold}.}
\label{tab: real_world}
\tablestyle{8pt}{0.85} 
\begin{tabular}{lcccccccccccc}
\toprule
 & \multicolumn{4}{c}{Vstar Bench} & \multicolumn{4}{c}{HR-Bench 4K} & \multicolumn{4}{c}{HR-Bench 8K} \\
 & \multicolumn{2}{c}{Attr} & \multicolumn{2}{c}{Spatial} & \multicolumn{2}{c}{FSP} & \multicolumn{2}{c}{FCP} & \multicolumn{2}{c}{FSP} & \multicolumn{2}{c}{FCP} \\
 \cmidrule(lr){2-3} \cmidrule(lr){4-5} \cmidrule(lr){6-7} \cmidrule(lr){8-9} \cmidrule(lr){10-11} \cmidrule(lr){12-13}
 Method & ACC & TSR & ACC & TSR & ACC & TSR & ACC & TSR & ACC & TSR & ACC & TSR \\
\midrule
GPT-4o & 72.2 & - & 60.5 & - & 66.8 & - & 63.3 & - & 60.8 & - & 58.5 & - \\
Thyme & 83.5 & - & 80.3 & - & 91.0 & - & 63.0 & - & 86.5 & - & 57.5 & - \\
DeepEyes & 91.3 & - & 88.2 & - & 91.3 & - & 59.0 & - & 86.5 & - & 58.5 & - \\
\midrule
\addlinespace[2pt]
\rowcolor{gray!10} \multicolumn{13}{c}{\textit{Qwen3-VL Thinking}} \\
\addlinespace[2pt]
Base & 78.3 & - & 73.7 & - & 83.0 & - & 57.3 & - & 78.3 & - & 56.0 \\
SC & 80.0 & - & 77.6 & - & 89.5 & - & 61.0 & - & 86.5 & - & 61.0 & - \\
ASC & 80.0 & 51.8 & 79.0 & 42.8 & 89.5 & 51.9 & 61.8 & 29.3 & 86.5 & 50.4 & 60.8 & 28.6 \\
ESC & 80.0 & 35.9 & 77.6 & 22.0 & 89.5 & 40.5 & 61.0 & 18.9 & 86.5 & 35.7 & 61.0 & 18.5 \\
CISC & 82.6 & 41.1 & 69.7 & 29.9 & 88.0 & 32.3 & 59.0 & 42.9 & 87.8 & 42.6 & 56.0 & 38.8 \\
Deepconf & 81.7 & 47.6 & 76.3 & 52.8 & 90.5 & 57.3 & 62.5 & 48.9 & 87.0 & 46.9 & 61.8 & 51.1 \\
Self-Cer. & 82.6 & 42.9 & 68.4 & 33.1 & 89.5 & 47.6 & 59.0 & 45.4 & 88.0 & 43.8 & 59.0 & 43.5 \\
\rowcolor{pink!30}Ours & \textbf{85.2} & \textbf{61.1} & \textbf{81.6} & \textbf{61.2} & \textbf{91.3} & \textbf{61.9} & \textbf{65.8} & \textbf{50.5} & \textbf{88.8} & \textbf{59.3} & \textbf{62.3} & \textbf{52.3} \\
\midrule
\addlinespace[2pt]
\rowcolor{gray!10} \multicolumn{13}{c}{\textit{Qwen3-VL Instruct}} \\
\addlinespace[2pt]
Base & 90.4 & - & 86.8 & - & 93.8 & - & 72.0 & - & 90.8 & - & 71.3 \\
SC & 92.2 & - & 89.5 & - & 95.3 & - & 77.0 & - & 93.0 & - & 74.8 & - \\
ASC & 93.0 & 59.4 & 89.5 & 56.8 & 95.8 & 59.8 & 76.8 & 39.9 & 93.0 & 54.6 & 74.8 & 38.2 \\
ESC & 92.2 & 56.1 & 89.5 & 50.9 & 95.3 & 60.7 & 77.0 & 23.1 & 93.0 & 51.7 & 74.8 & 25.6 \\
CISC & 93.9 & 55.8 & 89.5 & 56.6 & 95.5 & 57.7 & 75.8 & 41.2 & 91.0 & 53.6 & 75.8 & 30.0 \\
Deepconf & 93.9 & 57.3 & \textbf{90.8} & 55.0 & 96.0 & 60.3 & 76.8 & 38.3 & 93.5 & 58.9 & 75.5 & 32.1 \\
Self-Cer. & 92.2 & 59.6 & 89.5 & 57.0 & 95.3 & 62.5 & 75.8 & 43.2 & 88.8 & 57.1 & 75.3 & 30.8 \\
\rowcolor{pink!30}Ours & \textbf{94.8} & \textbf{60.4} & \textbf{90.8} & \textbf{58.5} & \textbf{96.5} & \textbf{66.2} & \textbf{78.3} & \textbf{46.4} & \textbf{94.8} & \textbf{60.3} & \textbf{77.0} & \textbf{45.8} \\
\bottomrule
\end{tabular}

\vspace{1.5em} 

\begin{minipage}[t]{0.51\textwidth}
    \centering
    \tablestyle{6.5pt}{0.85} 
    \caption{Results on TWI-oriented TreeBench under online setting.}
    \label{table: twi_oriented}
    \begin{tabular}{lcccccc}
    \toprule
     & \multicolumn{3}{c}{Reasoning} & \multicolumn{3}{c}{Perception} \\
     \cmidrule(lr){2-4} \cmidrule(lr){5-7}
     Method & ACC & mIoU & TSR & ACC & mIoU & TSR \\
    \midrule
    \addlinespace[2pt]
    \rowcolor{gray!10} \multicolumn{7}{c}{\textit{Qwen3-VL Thinking}} \\
    \addlinespace[2pt]
    Base & 32.8 & - & - & 63.1 & - & - \\
    SC & 30.9 & 37.8 & - & 67.1 & 32.5 & - \\
    ASC & 30.9 & 37.7 & 32.1 & 67.1 & 32.1 & 40.1 \\
    ESC & 30.9 & 37.7 & 10.7 & 67.1 & 33.4 & 28.0 \\
    CISC & 27.7 & 36.5 & 48.9 & 67.1 & 32.9 & 49.7 \\
    Deepconf & 35.6 & 39.3 & 51.7 & 65.8 & 31.9 & 53.5 \\
    Self-Cer. & 30.9 & 37.6 & 32.3 & 64.4 & 33.1 & 38.9 \\
    \rowcolor{pink!30} Ours & \textbf{37.1} & \textbf{40.8} & \textbf{59.4} & \textbf{67.8} & \textbf{35.1} & \textbf{60.1} \\
    \midrule
    \addlinespace[2pt]
    \rowcolor{gray!10} \multicolumn{7}{c}{\textit{Qwen3-VL Instruct}} \\
    \addlinespace[2pt]
    Base & 34.0 & - & - & 64.4 & - & - \\
    SC & 34.0 & 43.4 & - & 68.4 & 38.6 & - \\
    ASC & 34.0 & 43.6 & 41.2 & 68.4 & 38.9 & 47.9 \\
    ESC & 34.0 & 43.5 & 26.8 & 68.4 & 38.7 & 40.9 \\
    CISC & 34.4 & 43.5 & 38.6 & 67.8 & 37.9 & 43.2 \\
    Deepconf & 34.4 & 43.3 & 33.8 & 68.4 & 38.3 & 43.4 \\
    Self-Cer. & 34.4 & 43.4 & 42.3 & 68.4 & 38.9 & 48.5 \\
    \rowcolor{pink!30} Ours & \textbf{34.8} & \textbf{45.8} & \textbf{42.8} & \textbf{70.5} & \textbf{40.8} & \textbf{50.7} \\
    \bottomrule
    \end{tabular}
\end{minipage}
\hfill 
\begin{minipage}[t]{0.46\textwidth}
    \centering
    \tablestyle{8pt}{0.85}
    \caption{Results on reasoning benchmarks under online setting.}
    \label{table: multimodal_reasoning}
    \begin{tabular}{lcccc}
    \toprule
     & \multicolumn{2}{c}{MathVision} & \multicolumn{2}{c}{LogicVista} \\
     \cmidrule(lr){2-3} \cmidrule(lr){4-5}
    Method & ACC & TSR & ACC & TSR \\
    \midrule
    \addlinespace[2pt]
    \rowcolor{gray!10} \multicolumn{5}{c}{\textit{Qwen3-VL Thinking}} \\
    \addlinespace[2pt]
    Base & 18.7 & - & 44.3 & - \\
    SC & 19.9 & - & 46.4 & - \\
    ASC & 19.9 & 8.6 & 46.4 & 21.2 \\
    ESC & 19.9 & 5.5 & 44.3 & 18.9 \\
    CISC & 20.1 & 16.1 & 49.8 & 30.8 \\
    Deepconf & 21.2 & 33.3 & 51.8 & 34.5 \\
    Self-Cer. & 18.0 & 15.1 & 49.3 & 31.7 \\
    \rowcolor{pink!30} Ours & \textbf{23.2} & \textbf{34.6} & \textbf{61.7} & \textbf{50.5} \\
    \midrule
    \addlinespace[2pt]
    \rowcolor{gray!10} \multicolumn{5}{c}{\textit{Qwen3-VL Instruct}} \\
    \addlinespace[2pt]
    Base & 21.4 & - & 47.9 & - \\
    SC & 24.0 & - & 54.6 & - \\
    ASC & 24.0 & 17.2 & 54.1 & 27.6 \\
    ESC & 24.0 & 3.8 & 54.6 & 7.9 \\
    CISC & 22.4 & 8.1 & 53.4 & 14.1 \\
    Deepconf & 22.7 & 35.7 & 56.0 & 32.1 \\
    Self-Cer. & 24.0 & 7.7 & 54.8 & 15.1 \\
    \rowcolor{pink!30} Ours & \textbf{26.1} & \textbf{37.6} & \textbf{58.5} & \textbf{34.7} \\
    \bottomrule
    \end{tabular}
\end{minipage}

\end{table*}

\subsection{Reliable Voting}
\label{sec: reliable_voting}
Although the dual-stage filtering in Eq.~\ref{eq: dual-stage filtering} could exclude unreliable traces with NT to a great extent, it is still inadequate for determining trustworthy answer among the remaining ones.
One straightforward solution is to aggregate the answers via majority voting~\cite{sc}, \textit{i.e.}, select the most common answer.
However, such a simple approach treats each trace and the corresponding answer equally, regardless of their quality.
To remedy this, we propose a novel reliable voting module to derive the trustworthy answer $\hat{A}$ as follows,
\begin{equation}
\begin{gathered}
    \hat{A}=\arg\max_{a}\sum_{t \in T_{\text{rel}}}
C_t\cdot\mathbb{I}\big[A_t=a\big],\\
C_t=exp\left(\Delta_t/\left(\left|w_t\right|\cdot \tau \right)\right),
\end{gathered}
\label{eq: reliable_voting}
\end{equation}
where $w_t$ denotes the trace-wise reliability, $\Delta_t$ denotes the reliability leap, $\tau$ is the temperature, $\mathbb{I}[\cdot]$ is an indicator function evaluating to $1$ \textit{i.f.f.} the condition is satisfied.
In particular, the confidence weight $C_t$ is determined by the dedicated two-fold principle.
Inspired by the observations in Fig.~\ref{fig: fig1}(c) and Fig.~\ref{fig: observation_3}, the traces with correct answers always exhibit higher dual-stage reliability and reliability leap, which could serve as two principles to determine the trustworthy answers.
Formally, we give the following definition of the trace reliability.
\begin{definition}[Trace Reliability]
    For a given thinking trace $t$, the trace-wise reliability is defined as
    \begin{equation}
        w_t=\sum_{s\in\{m,r\}} w(t^{s}).
    \end{equation}
\end{definition}
Such a formulation encourages traces with reliable mining and reasoning stages to yield higher confidence.
Moreover, we introduce the definition of reliability leap as follows.
\begin{definition}[Reliability Leap]
    For a given two-stage thinking trace $t=\{t^m,t^r\}$, the reliability leap is defined as
    \begin{equation}
        \Delta_t=\max\big(w(t^{r})-w(t^{m}),\,0\big),
    \end{equation}
    where $\max (\cdot,0)$ ensures the reliability leap is non-negative.
\end{definition}
The design of $\Delta_t$ could mine the potential correct answer by prioritizing traces with substantial reliability leap.
In other words, beyond traces that are reliable in both stages, some traces may exhibit low reliability during the mining stage due to ambiguous multimodal understanding, yet still generate confident textual CoT and ultimately arrive at the correct answer after incorporating appropriate visual cues.

\begin{figure*}[t]
    \centering
    \includegraphics[width=0.85\linewidth]{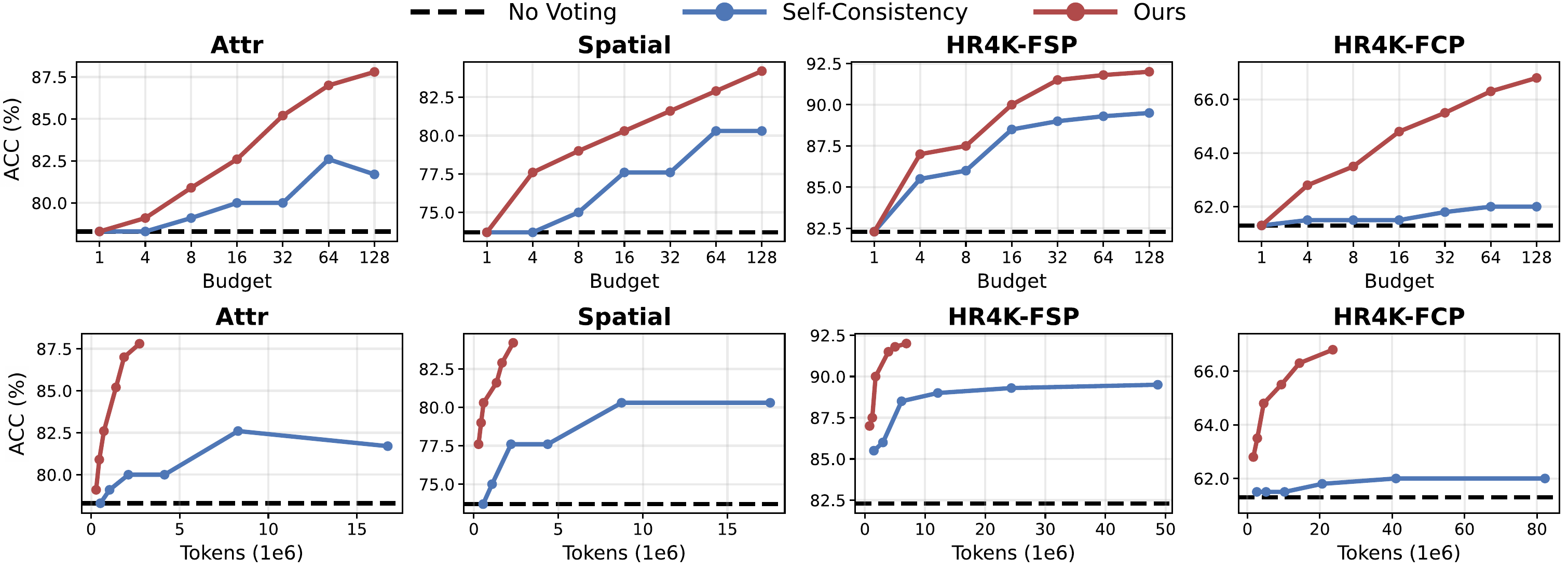}
    \caption{Test-time Scaling. The first row indicates the accuracy as the budget increases.
    The second row illustrates the relationship between accuracy and generated tokens under various test-time costs.}
    \label{fig: budet}
\end{figure*}

\begin{figure}[t]
    \centering
    \includegraphics[width=0.95\linewidth]{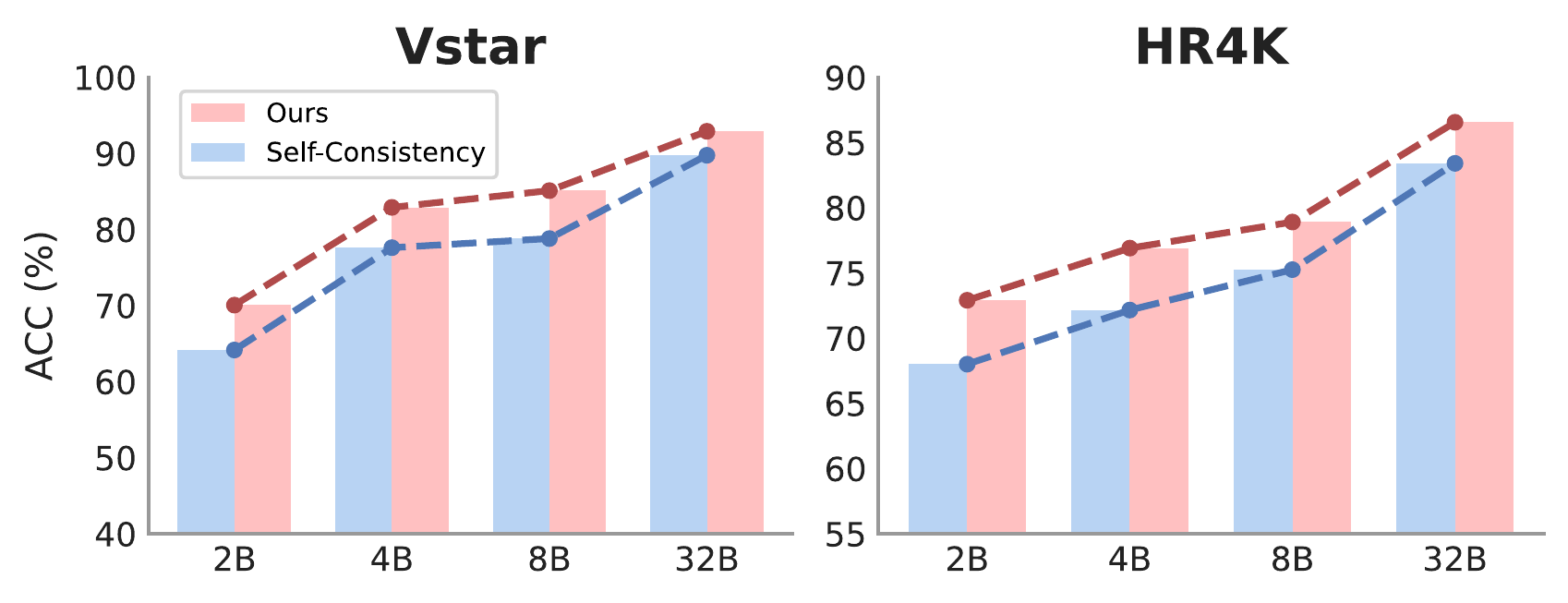}
    \caption{Scaling law across model sizes.}
    \label{fig: scale}
\end{figure}

\begin{table}[t]
\centering
\tablestyle{8pt}{0.9} 
\caption{Ablation study of the designs in RTWI, where ``w/o DF'' denotes using only the consensus-based filtering, ``w/o RV'' denotes no usage of reliable voting.}
\label{tab: ablation_variants}
\begin{tabular}{lcccc}
\toprule
& \multicolumn{2}{c}{Vstar Bench} & \multicolumn{2}{c}{HR-Bench 4K} \\
\cmidrule(lr){2-3} \cmidrule(lr){4-5}
Variants & ACC & TSR & ACC & TSR \\
\midrule
SC & 78.8 & - & 75.3 & - \\
w/o DF & 80.8 & 35.6 & 77.1 & 38.4 \\
w/o RV & 81.9 & 61.2 & 77.9 & 56.2 \\
w/o $w_t$ & 82.3 & 58.3 & 77.9 & 50.2 \\
w/o $\Delta_t$ & 82.3 & 57.5 & 78.0 & 54.1 \\
\midrule
Default & 83.4 & 61.2 & 78.6 & 56.2 \\
\bottomrule
\end{tabular}
\end{table}

\begin{figure}[t]
    \centering
    \includegraphics[width=0.95\linewidth]{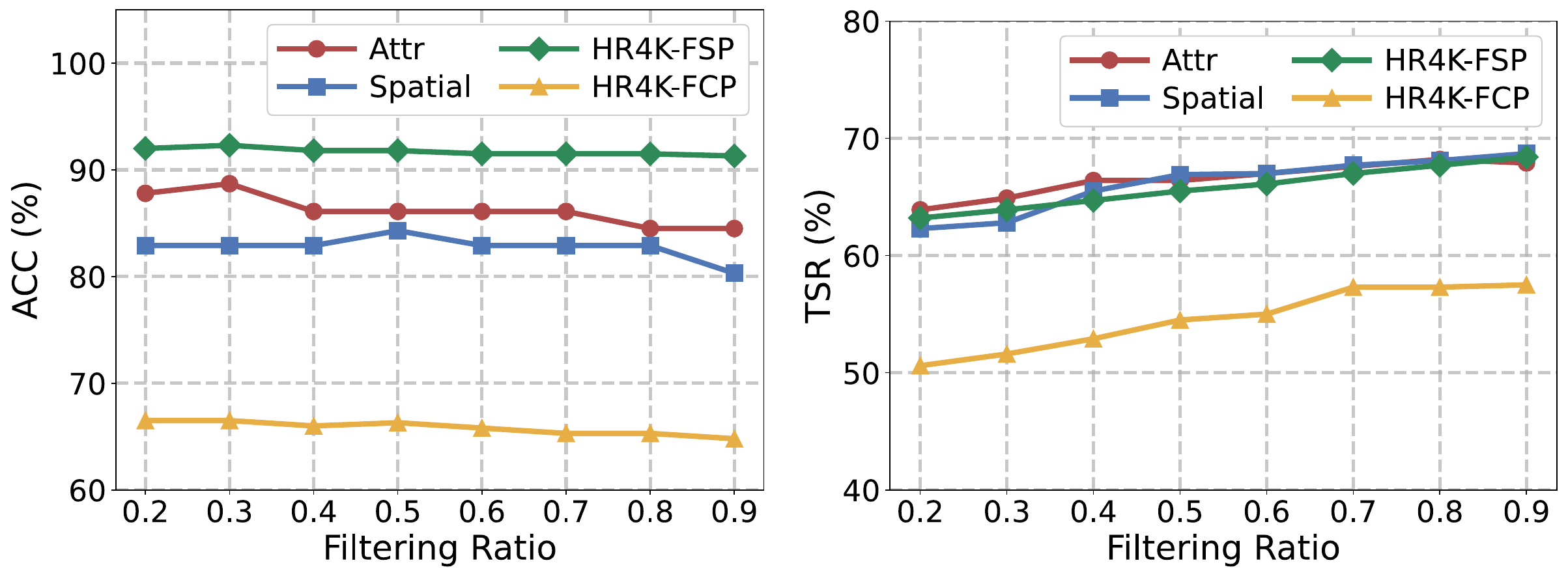}
    \caption{The parameter analysis of filtering ratio $\alpha$ in Eq~\ref{eq: dual-level thresholds}.}
    \label{fig: ablation_threshold}
\end{figure}

\section{Experiments}
In this section, we conduct expensive experiments on widely-used TWI benchmarks to verify the effectiveness of the proposed RTWI. Due to space limitation, we present more experimental details and results in Appendix~\ref{appendix: more_experimental_results}.
\subsection{Experiment Configurations}
\textbf{Settings.} Following~\cite{esc,deepconf}, we adopt two widely-used evaluation protocols, namely, online and offline settings. 
In the online setting, we perform on-the-fly token generation with up to $32$ thinking traces per question, evaluating whether the methods could improve both reasoning accuracy and computational efficiency.
In the offline setting, $32$ thinking traces are pre-generated for each problem, and the key challenge is deriving trustworthy answer from multiple traces.
Note that the number of warm-up traces $|T_{\text{sel}}|$ in the offline and online setting are set to $32$ and $8$, respectively.

\textbf{Benchmarks.} To comprehensively evaluate the effectiveness of RTWI, we evaluate our method on real-world high-resolution benchmarks, including Vstar Bench~\cite{vstar}, HR-Bench~\cite{hrbench} at 4K and 8K resolutions, TWI-oriented TreeBench~\cite{treebench}, and multimodal reasoning benchmarks including MathVision~\cite{mathvision} and LogicVista~\cite{logicvista}.

\textbf{Implementation Details.} 
We evaluate our proposed RTWI using the SOTA TWI model Qwen3-VL~\cite{Qwen3-VL} and conduct comprehensive experiments on two variants, \textit{i.e.}, Qwen3-VL-Thinking and Qwen3-VL-Instruct.
Regarding hyperparameter settings, the filtering ratio $\alpha$ in Eq.~\ref{eq: dual-level thresholds} is set to $0.4$, while the temperature in Eq.~\ref{eq: reliable_voting} is fixed to $0.1$ for the Thinking variant and $1.0$ for the Instruct one.

\subsection{Comparisons with State-Of-The-Arts}
We compare RTWI with three SOTA TWI baselines, including GPT-4o~\cite{gpt4}, Thyme~\cite{thyme}, DeepEyes~\cite{deepeyes} and six SOTA TTS scaling methods on different benchmarks under the online setting, including the vanilla self-consistency method~\cite{sc}, early stopping methods~\cite{asc,esc}, robust reasoning method~\cite{deepconf,self-certainty,cisc}. 
Specifically, in the online setting, following DeepConf~\cite{deepconf}, RTWI adopts an early-stopping strategy once the generated traces reach a consensus threshold $\beta=0.9$, thus reducing the trace generation budget.
Besides, for methods not specially designed for the online setting, we employ the same early-stopping strategy to enable their evaluation under online setting.
For fair comparisons, we pre-generate $32$ complete thinking traces for each question, and use them for evaluations among various baselines.
For more results under offline setting, please refer to Appendix~\ref{appendix: offline_evaluation}.

From the results in Table~\ref{tab: real_world}-\ref{table: multimodal_reasoning}, one could have the following observations and conclusions:
i) existing methods yield marginal performance gains over the vanilla self-consistency method, which could be attributed to the inability to identify and mitigate NT. 
In contrast, RTWI achieves robust filtering and voting against NT, thus significantly outperforming all the baselines across various model variants and benchmarks;
ii) compared with existing early stopping and robust reasoning baselines, RTWI significantly avoids unnecessary token generation and achieves notable gains in computational efficiency, which might be attributed to our advanced reliability estimation of traces;
iii) as shown in Table~\ref{table: twi_oriented}, RTWI achieves higher mean IOU (see more details in Appendix~\ref{sec: mIoU}) between the predicted and ground-truth visual cues of filtered traces, demonstrating that RTWI improves the final accuracy by excluding traces with unreliable visual cues.

\begin{table}[t]
\centering
\tablestyle{4pt}{1.0}
\caption{Analytic study on the ratio of high-entropy tokens in the tool-invocation and vanilla periods.}
\label{tab: high_entropy_token_ablation}
\begin{tabular}{lcc}
\toprule
Period & Vstar Bench & HR-Bench 4K \\
\midrule
Tool Invocation & 68.6 & 66.0 \\
Vanilla & 31.4 & 34.0 \\
\bottomrule
\end{tabular}
\end{table}

\begin{table}[t]
\centering
\tablestyle{4pt}{0.9} 
\caption{Analytic study about visual cue consistency on TreeBench.}
\label{tab: consistency_ablation}
\begin{tabular}{lcccc}
\toprule
& \multicolumn{2}{c}{Vstar Bench} & \multicolumn{2}{c}{HR-Bench 4K} \\
\cmidrule(lr){2-3} \cmidrule(lr){4-5}
Method & Consistency & ACC & Consistency & ACC \\
\midrule
SC        & 43.0 & 78.8 & 48.4 & 75.3 \\
CISC      & 44.5 & 76.2 & 49.3 & 73.5 \\
Deepconf  & 44.9 & 79.0 & 50.0 & 76.5 \\
Self-Cer. & 44.1 & 75.5 & 49.0 & 74.3 \\
\rowcolor{pink!30} Ours & \textbf{47.6} & \textbf{83.4} & \textbf{51.7} & \textbf{78.6} \\
\bottomrule
\end{tabular}
\end{table}

\begin{figure*}[t]
    \centering
    \includegraphics[width=0.9\linewidth]{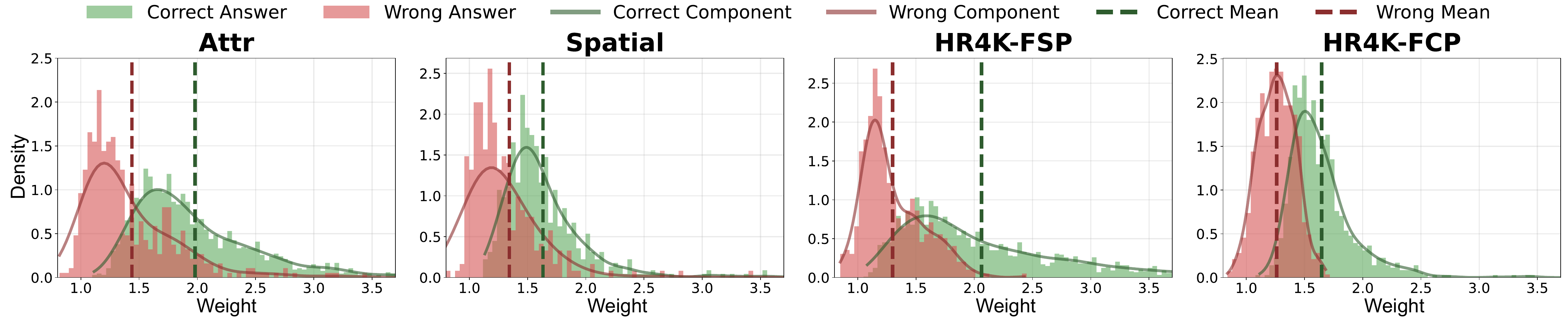}
    \caption{Analytical study of confidence distributions for correct and incorrect thinking traces.}
    \label{fig: gmm}
\end{figure*}

\subsection{Ablation and Analytic Study}
In this section, we carry out a series of ablation studies and analytic experiments to investigate the effectiveness of RTWI.
Unless otherwise stated, all the experiments are conducted under the online setting using the Qwen3-VL-8B-Thinking.

\textbf{Test-time Scaling.} 
To comprehensively investigate the effectiveness of RTWI, we conduct experiments using various test-time budgets and model sizes.
From the results in Fig.~\ref{fig: budet}, one could have the following conclusions:
i) RTWI yields significant performance gains over the vanilla self-consistency as the budget scales, demonstrating a favorable scaling behavior with respect to computational budget;
ii) RTWI exhibits favorable efficiency-accuracy trade-offs as generated token scales.
Moreover, we evaluate the proposed RTWI using Qwen3-VL Thinking models with 2B, 4B, 8B, and 32B parameters.
As illustrated in Fig.~\ref{fig: scale}, RTWI achieves consistent performance improvements as the model size scales, which aligns with the expected scaling law~\cite{scaling_law}.
In particular, the performance gains are more pronounced for models with limited reasoning capacity, \textit{e.g.}, Qwen3-VL-2B-Thinking.

\textbf{Ablation studies.} 
To verify the importance of each design, we investigate the variants of the RTWI in Table~\ref{tab: ablation_variants}, where one could have the following conclusions.
On the one hand, both the proposed Dual-stage Filtering (DF) module and Reliable Voting (RV) module could select and prioritize reliable traces and thus boost the reasoning accuracy.
On the other hand, the two-fold reliability principles could complement each other and thus facilitate accurate answer derivation and unreliable trace exclusion.
As a result, RTWI could achieve better reasoning performance while saving more test-time compute.
Besides, we carry out parameter analysis of the filtering ratio $\alpha$ in Eq.~\ref{eq: dual-level thresholds}.
As depicted in Fig.~\ref{fig: ablation_threshold}, RTWI demonstrates stable performance with low token consumption when $\alpha$ falls within the range of $[0.3, 0.5]$.

\textbf{Analytic Study on Reliability Estimation.} 
As pointed out in Section~\ref{sec: reliability_estimation}, we estimate the reliability of visual cues by considering high-entropy tokens in textual CoT.
To verify this claim, we conduct an analytic study on the distribution of Top-$k$ high-entropy tokens in the mining stage.
As shown in Table~\ref{tab: high_entropy_token_ablation}, the ratio of high-entropy tokens in tool invocation is consistently higher than that in vanilla generation, indicating that RTWI focuses on tool-invocation tokens and thus effectively estimates the reliability of visual cues.

\textbf{Analytic Study on Dual-stage Filtering.}
To demonstrate the effectiveness of the dual-stage filtering module, we conduct an analytical study on the visual cue consistency among the filtered traces $T_{\text{ref}}$, see more details in Appendix~\ref{appendix: visual_cue_consistency}.
As shown in Table~\ref{tab: consistency_ablation}, RTWI achieves the highest consistency compared with the most competitive methods in Table~\ref{tab: real_world}-\ref{table: multimodal_reasoning}, which indicates that RTWI improves reasoning accuracy by prioritizing its focus on similar task-relevant regions.

\textbf{Analytic Study on Reliable Voting.} 
To verify the effectiveness of the reliable voting module, we visualize the confidence distribution of traces with correct or incorrect answers.
From the results in Fig.~\ref{fig: gmm}, traces with correct answers generally exhibit higher confidence than those with incorrect answers, indicating that confidence serves as an effective indicator for identifying NT.

\section{Conclusion}
In this paper, we study a new problem in TWI, i.e., Noisy Thinking, which refers to incorrect cue mining and answer reasoning stages.
To solve this problem, the proposed RTWI employs a unified text-centric reliability estimation mechanism to identify NT and accordingly mitigate the negative impact of NT by reliability-aware filtering and voting.
Extensive experiments on real-world benchmarks under offline and online settings verify the effectiveness of RTWI against NT.
In the future, we plan to explore the reliable TTS paradigm in a broader range of agent applications, advancing agentic systems toward reliable reasoning.

\bibliography{reference}

@string{AAAI = "AAAI"}

@string{CVPR = "CVPR"}

@string{ICLR = "ICLR"}

@string{NEURIPS = "NeurIPS"}

@inproceedings{cot1,
  title={Chain-of-thought prompting elicits reasoning in large language models},
  author={Wei, Jason and Wang, Xuezhi and Schuurmans, Dale and Bosma, Maarten and Xia, Fei and Chi, Ed and Le, Quoc V and Zhou, Denny and others},
  booktitle={NeurIPS},
  year={2022}
}

@article{cot2,
  title={Automatic chain of thought prompting in large language models},
  author={Zhang, Zhuosheng and Zhang, Aston and Li, Mu and Smola, Alex},
  journal={arXiv preprint arXiv:2210.03493},
  year={2022}
}

@article{llm1,
  title={A comprehensive overview of large language models},
  author={Naveed, Humza and Khan, Asad Ullah and Qiu, Shi and Saqib, Muhammad and Anwar, Saeed and Usman, Muhammad and Akhtar, Naveed and Barnes, Nick and Mian, Ajmal},
  journal={ACM Transactions on Intelligent Systems and Technology},
  year={2025},
  publisher={ACM New York, NY}
}

@article{llm2,
  title={A survey of large language models},
  author={Zhao, Wayne Xin and Zhou, Kun and Li, Junyi and Tang, Tianyi and Wang, Xiaolei and Hou, Yupeng and Min, Yingqian and Zhang, Beichen and Zhang, Junjie and Dong, Zican and others},
  journal={arXiv preprint arXiv:2303.18223},
  volume={1},
  number={2},
  year={2023}
}

@article{mllm1,
  title={Multimodal chain-of-thought reasoning in language models},
  author={Zhang, Zhuosheng and Zhang, Aston and Li, Mu and Zhao, Hai and Karypis, George and Smola, Alex},
  journal={arXiv preprint arXiv:2302.00923},
  year={2023}
}

@inproceedings{mllm2,
  title={Ddcot: Duty-distinct chain-of-thought prompting for multimodal reasoning in language models},
  author={Zheng, Ge and Yang, Bin and Tang, Jiajin and Zhou, Hong-Yu and Yang, Sibei},
  booktitle={NeurIPS},
  year={2023}
}

@article{sc,
  title={Self-consistency improves chain of thought reasoning in language models},
  author={Wang, Xuezhi and Wei, Jason and Schuurmans, Dale and Le, Quoc and Chi, Ed and Narang, Sharan and Chowdhery, Aakanksha and Zhou, Denny},
  journal={arXiv preprint arXiv:2203.11171},
  year={2022}
}

@inproceedings{asc,
  title={Let’s Sample Step by Step: Adaptive-Consistency for Efficient Reasoning and Coding with LLMs},
  author={Aggarwal, Pranjal and Madaan, Aman and Yang, Yiming and others},
  booktitle={EMNLP},
  year={2023}
}

@inproceedings{esc,
  title={Escape Sky-high Cost: Early-stopping Self-Consistency for Multi-step Reasoning},
  author={Li, Yiwei and Yuan, Peiwen and Feng, Shaoxiong and Pan, Boyuan and Wang, Xinglin and Sun, Bin and Wang, Heda and Li, Kan},
  booktitle={ICLR},
  year={2024}
}

@article{cisc,
  title={Confidence improves self-consistency in llms},
  author={Taubenfeld, Amir and Sheffer, Tom and Ofek, Eran and Feder, Amir and Goldstein, Ariel and Gekhman, Zorik and Yona, Gal},
  journal={arXiv preprint arXiv:2502.06233},
  year={2025}
}

@article{self-certainty,
  title={Scalable best-of-n selection for large language models via self-certainty},
  author={Kang, Zhewei and Zhao, Xuandong and Song, Dawn},
  journal={arXiv preprint arXiv:2502.18581},
  year={2025}
}

@article{deepconf,
  title={Deep think with confidence},
  author={Fu, Yichao and Wang, Xuewei and Tian, Yuandong and Zhao, Jiawei},
  journal={arXiv preprint arXiv:2508.15260},
  year={2025}
}

@article{Qwen3-VL,
      title={Qwen3-VL Technical Report}, 
      author={Shuai Bai and Yuxuan Cai and Ruizhe Chen and Keqin Chen and Xionghui Chen and Zesen Cheng and Lianghao Deng and Wei Ding and Chang Gao and Chunjiang Ge and Wenbin Ge and Zhifang Guo and Qidong Huang and Jie Huang and Fei Huang and Binyuan Hui and Shutong Jiang and Zhaohai Li and Mingsheng Li and Mei Li and Kaixin Li and Zicheng Lin and Junyang Lin and Xuejing Liu and Jiawei Liu and Chenglong Liu and Yang Liu and Dayiheng Liu and Shixuan Liu and Dunjie Lu and Ruilin Luo and Chenxu Lv and Rui Men and Lingchen Meng and Xuancheng Ren and Xingzhang Ren and Sibo Song and Yuchong Sun and Jun Tang and Jianhong Tu and Jianqiang Wan and Peng Wang and Pengfei Wang and Qiuyue Wang and Yuxuan Wang and Tianbao Xie and Yiheng Xu and Haiyang Xu and Jin Xu and Zhibo Yang and Mingkun Yang and Jianxin Yang and An Yang and Bowen Yu and Fei Zhang and Hang Zhang and Xi Zhang and Bo Zheng and Humen Zhong and Jingren Zhou and Fan Zhou and Jing Zhou and Yuanzhi Zhu and Ke Zhu},
	  journal={arXiv preprint arXiv:2511.21631},
      year={2025}
}

@article{top10,
  title={The curious case of neural text degeneration},
  author={Holtzman, Ari and Buys, Jan and Du, Li and Forbes, Maxwell and Choi, Yejin},
  journal={arXiv preprint arXiv:1904.09751},
  year={2019}
}

@article{treebench,
  title={Traceable evidence enhanced visual grounded reasoning: Evaluation and methodology},
  author={Wang, Haochen and Li, Xiangtai and Huang, Zilong and Wang, Anran and Wang, Jiacong and Zhang, Tao and Zheng, Jiani and Bai, Sule and Kang, Zijian and Feng, Jiashi and others},
  journal={arXiv preprint arXiv:2507.07999},
  year={2025}
}

@article{deepeyes,
  title={DeepEyes: Incentivizing" Thinking with Images" via Reinforcement Learning},
  author={Zheng, Ziwei and Yang, Michael and Hong, Jack and Zhao, Chenxiao and Xu, Guohai and Yang, Le and Shen, Chao and Yu, Xing},
  journal={arXiv preprint arXiv:2505.14362},
  year={2025}
}

@article{thyme,
  title={Thyme: Think beyond images},
  author={Zhang, Yi-Fan and Lu, Xingyu and Yin, Shukang and Fu, Chaoyou and Chen, Wei and Hu, Xiao and Wen, Bin and Jiang, Kaiyu and Liu, Changyi and Zhang, Tianke and others},
  journal={arXiv preprint arXiv:2508.11630},
  year={2025}
}

@article{minio3,
  title={Mini-o3: Scaling up reasoning patterns and interaction turns for visual search},
  author={Lai, Xin and Li, Junyi and Li, Wei and Liu, Tao and Li, Tianjian and Zhao, Hengshuang},
  journal={arXiv preprint arXiv:2509.07969},
  year={2025}
}

@article{gpt4,
  title={Gpt-4 technical report},
  author={Achiam, Josh and Adler, Steven and Agarwal, Sandhini and Ahmad, Lama and Akkaya, Ilge and Aleman, Florencia Leoni and Almeida, Diogo and Altenschmidt, Janko and Altman, Sam and Anadkat, Shyamal and others},
  journal={arXiv preprint arXiv:2303.08774},
  year={2023}
}

@misc{openaio3,
  author       = {{OpenAI}},
  title        = {Thinking with Images},
  year         = {2025},
  howpublished = {\url{https://openai.com/index/thinking-with-images/}}
}

@article{openaio1,
  title={Openai o1 system card},
  author={Jaech, Aaron and Kalai, Adam and Lerer, Adam and Richardson, Adam and El-Kishky, Ahmed and Low, Aiden and Helyar, Alec and Madry, Aleksander and Beutel, Alex and Carney, Alex and others},
  journal={arXiv preprint arXiv:2412.16720},
  year={2024}
}

@article{deepseekr1,
  title={Deepseek-r1: Incentivizing reasoning capability in llms via reinforcement learning},
  author={Guo, Daya and Yang, Dejian and Zhang, Haowei and Song, Junxiao and Zhang, Ruoyu and Xu, Runxin and Zhu, Qihao and Ma, Shirong and Wang, Peiyi and Bi, Xiao and others},
  journal={arXiv preprint arXiv:2501.12948},
  year={2025}
}

@article{qwen3,
  title={Qwen3 technical report},
  author={Yang, An and Li, Anfeng and Yang, Baosong and Zhang, Beichen and Hui, Binyuan and Zheng, Bo and Yu, Bowen and Gao, Chang and Huang, Chengen and Lv, Chenxu and others},
  journal={arXiv preprint arXiv:2505.09388},
  year={2025}
}

@misc{grok4,
  title       = {Grok 4 Model Card},
  author      = {{xAI}},
  year        = {2025},
  howpublished = {\url{https://data.x.ai/2025-08-20-grok-4-model-card.pdf}}
}

@inproceedings{zoomeye,
  title={Zoomeye: Enhancing multimodal llms with human-like zooming capabilities through tree-based image exploration},
  author={Shen, Haozhan and Zhao, Kangjia and Zhao, Tiancheng and Xu, Ruochen and Zhang, Zilun and Zhu, Mingwei and Yin, Jianwei},
  booktitle={EMNLP},
  year={2025}
}

@inproceedings{dyfo,
  title={Dyfo: A training-free dynamic focus visual search for enhancing lmms in fine-grained visual understanding},
  author={Li, Geng and Xu, Jinglin and Zhao, Yunzhen and Peng, Yuxin},
  booktitle={CVPR},
  year={2025}
}

@article{scaling_law,
  title={Scaling laws for neural language models},
  author={Kaplan, Jared and McCandlish, Sam and Henighan, Tom and Brown, Tom B and Chess, Benjamin and Child, Rewon and Gray, Scott and Radford, Alec and Wu, Jeffrey and Amodei, Dario},
  journal={arXiv preprint arXiv:2001.08361},
  year={2020}
}

@inproceedings{vstar,
  title={V?: Guided visual search as a core mechanism in multimodal llms},
  author={Wu, Penghao and Xie, Saining},
  booktitle={CVPR},
  year={2024}
}

@inproceedings{hrbench,
  title={Divide, conquer and combine: A training-free framework for high-resolution image perception in multimodal large language models},
  author={Wang, Wenbin and Ding, Liang and Zeng, Minyan and Zhou, Xiabin and Shen, Li and Luo, Yong and Yu, Wei and Tao, Dacheng},
  booktitle={AAAI},
  year={2025}
}

@inproceedings{mathvision,
  title={Measuring multimodal mathematical reasoning with math-vision dataset},
  author={Wang, Ke and Pan, Junting and Shi, Weikang and Lu, Zimu and Ren, Houxing and Zhou, Aojun and Zhan, Mingjie and Li, Hongsheng},
  booktitle={NeurIPS},
  year={2024}
}

@article{logicvista,
  title={Logicvista: Multimodal llm logical reasoning benchmark in visual contexts},
  author={Xiao, Yijia and Sun, Edward and Liu, Tianyu and Wang, Wei},
  journal={arXiv preprint arXiv:2407.04973},
  year={2024}
}

@article{semantic_entropy,
  title={Detecting hallucinations in large language models using semantic entropy},
  author={Farquhar, Sebastian and Kossen, Jannik and Kuhn, Lorenz and Gal, Yarin},
  journal={Nature},
  year={2024},
  publisher={Nature Publishing Group UK London}
}

@article{tts1,
  title={A Survey on Test-Time Scaling in Large Language Models: What, How, Where, and How Well?},
  author={Zhang, Qiyuan and Lyu, Fuyuan and Sun, Zexu and Wang, Lei and Zhang, Weixu and Hua, Wenyue and Wu, Haolun and Guo, Zhihan and Wang, Yufei and Muennighoff, Niklas and others},
  journal={arXiv preprint arXiv:2503.24235},
  year={2025}
}

@inproceedings{tts2,
  title={s1: Simple test-time scaling},
  author={Muennighoff, Niklas and Yang, Zitong and Shi, Weijia and Li, Xiang Lisa and Fei-Fei, Li and Hajishirzi, Hannaneh and Zettlemoyer, Luke and Liang, Percy and Cand{\`e}s, Emmanuel and Hashimoto, Tatsunori B},
  booktitle={EMNLP},
  year={2025}
}

@article{search-tts1,
  title={Mur: Momentum uncertainty guided reasoning for large language models},
  author={Yan, Hang and Xu, Fangzhi and Xu, Rongman and Li, Yifei and Zhang, Jian and Luo, Haoran and Wu, Xiaobao and Tuan, Luu Anh and Zhao, Haiteng and Lin, Qika and others},
  journal={arXiv preprint arXiv:2507.14958},
  year={2025}
}

@inproceedings{search-tts2,
  title={Self-evaluation guided beam search for reasoning},
  author={Xie, Yuxi and Kawaguchi, Kenji and Zhao, Yiran and Zhao, James Xu and Kan, Min-Yen and He, Junxian and Xie, Michael},
  booktitle={NeurIPS},
  year={2023}
}

@article{DRIM,
  title={Deep But Reliable: Advancing Multi-turn Reasoning for Thinking with Images},
  author={Yang, Wenhao and Xia, Yu and Huang, Jinlong and Lu, Shiyin and Chen, Qing-Guo and Xu, Zhao and Luo, Weihua and Zhang, Kaifu and Wan, Yuanyu and Zhang, Lijun},
  journal={arXiv preprint arXiv:2512.17306},
  year={2025}
}
\bibliographystyle{icml2026}

\newpage

\appendix
\onecolumn






\newpage

\section{More Implementation Details}
In this section, we provide more implementation details about the reliability estimation, online filtering, mIoU metric, and generation hyperparameter.

\subsection{More Details about Reliability Estimation for Single-Stage Traces}
\label{sec: single-stage reliability}
As discussed in~\cite{deepeyes,thyme}, MLLM would adaptively decide whether to adopt external tools for visual cue extraction based on the difficulty of the question.
Accordingly, not all generated traces involve tool usage, making it infeasible to derive confidence weights for all traces using Eq.~\ref{eq: reliable_voting}.
To remedy this, for a given single-stage trace $t=\{t^r\}$, we estimate its confidence weight as follows,
\begin{equation}
    C_t=exp\left(\left(\eta^{r}-\eta^{m}\right)/\left(|w_t|\cdot \tau\right)\right),
\end{equation}
where $w_t=2w(t^r)$.
Such a formulation not only assigns limited confidence weight to unreliable single-stage traces, but also remains consistent with the weighting mechanism in Eq.~\ref{eq: reliable_voting}.

\subsection{More Details about Dual-stage Filtering under Online Setting}
\label{appenidx: online_filtering}
In the online setting, TWI generates multiple turns of interleaved image-text CoT and then the final turn with textual CoT.
However, during online generation, the stage of the current turn is unknown. 
As a result, it is infeasible to employ the stage-specific thresholds to select reliable traces.
To remedy this, at each turn $i$, we dynamically estimate the reliability
of the currently generated textual CoT $\mathbf{x}_i^{\text{txt}}$, \textit{i.e.},
\begin{equation}
    w(\mathbf{x}_i^{\text{txt}})=-\frac{1}{k}\sum_{j \in \mathcal{K}\left(\mathcal{H}(\mathbf{x}_i^{\text{txt}})\right)} h_j ,
\end{equation}
where $\mathcal{H}(\mathbf{x}_i^{\text{txt}})$ denotes the set of token entropies at the $i$-th turn.
During online generation, we perform a preliminary filtering using the loose threshold $\eta^{m}$, which is motivated by the empirical observation that
$\eta^{m} < \eta^{r}$ as shown in Fig.~\ref{fig: observation_3}.
For a given thinking trace $t$ with multi-turn textual CoT
$\{\mathbf{x}_i^{\text{txt}}\}_{i=1}^{N}$, the trace is preserved during the online generation only if
\begin{equation}
    w(\mathbf{x}_i^{\text{txt}}) \ge \eta^{m}, \quad \forall i = 1, \ldots, N .
\end{equation}
After online filtering, the remaining traces are further subjected to the more stringent dual-stage filtering described in Eq.~\ref{eq: dual-stage filtering} to determine the final set of reliable traces.

\subsection{More Details about Reliability Estimation in Eq.~\ref{eq: stage_reliability}}
The number $k$ for reliability estimation in Eq.~\ref{eq: stage_reliability} is adaptively determined as
\begin{equation}
\arg\max_{k} \sum_{t \in T_{\text{sel}}} \big( w(t^{r}) - w(t^{m}) \big),
\end{equation}
which maximizes the reliability leap to facilitate the identification of reliable traces. Besides, following~\cite{top10}, token entropy is computed by employing the normalized distribution over the Top-$10$ token probabilities.

\subsection{More Details about the mIoU Metric}
\label{sec: mIoU}
The TWI-oriented TreeBench provides reference localizations of key visual cues, \textit{i.e.}, bounding boxes, which could be employed to evaluate the effectiveness of the proposed robust filtering mechanism.
Following~\cite{treebench}, we employ the mean Intersection-Over-Union (mIoU) metric to measure the alignment between predicted and ground-truth visual cues. 

Specifically, for a given multimodal question with ground-truth bounding boxes $\{b_k\}_{k=1}^{M}$, each filtered trace $t \in T_{\text{rel}}$ generate a set of predicted bounding boxes $\{\hat{b}^{t}_i\}_{i=1}^{N_t}$.
Accordingly, we define the trace-level IoU as
\begin{equation}
    \mathrm{IoU}(t)=\frac{1}{N_t}\sum_{i=1}^{N_t}\mathrm{IoU}\!\left(\{b_k\}_{k=1}^{M},\, \hat{b}^{t}_i\right),
\end{equation}
where $\mathrm{IoU}\!\left(\{b_k\}_{k=1}^{M},\, \hat{b}^{t}_i\right)=\max_{k} \ \mathrm{IoU}\!\left(b_k, \hat{b}^{t}_i\right)$.
After that, the overall mIoU score over the filtered trace set $T_{\text{rel}}$ is derived as
\begin{equation}
    \mathrm{mIoU}(T_{\text{rel}})=\frac{1}{|T{_\text{rel}}|}\sum_{t \in T_{\text{rel}}}\mathrm{IoU}(t).
\end{equation}
The formulation of mIoU reflects the intuition that each predicted bounding box should correspond to at least one ground-truth visual cue.

\subsection{More Details about Visual Cue Consistency}
\label{appendix: visual_cue_consistency}
In the manuscript, we conduct an analytical study on the visual cue consistency among the filtered traces. 
Specifically, given the filtered trace set $T_{\text{rel}}$, we define visual cue consistency as the average IoU between pairwise visual cues among the filtered traces, \textit{i.e.},
\begin{equation}
\mathrm{Cons}(T_{\text{rel}})
=
\frac{1}{
\displaystyle
\sum_{\substack{t, t' \in T_{\text{rel}} \\ t \neq t'}}
N_t \, N_{t'}
}\displaystyle
\sum_{\substack{t, t' \in T_{\text{rel}} \\ t \neq t'}}
\ \sum_{i=1}^{N_t}
\ \sum_{j=1}^{N_{t'}}
\mathrm{IoU}\!\left(\hat{b}_i^{t}, \hat{b}_j^{t'}\right).
\end{equation}
where $\{ \hat{b}_i^t \}_{i=1}^{N_t}$ denote the predicted bounding boxes produced by trace $t$.

\subsection{Generation Hyperparameters}
We list the per-model decoding hyperparameters used across all experiments. For each model, we fix the temperature, top-$p$, top-$k$, and the maximum generation length.
\begin{table}[ht]
\centering
\caption{Generation hyperparameters used in our experiments. Different models use different decoding settings.}
\label{tab:generation_hyperparams}
\tablestyle{8pt}{0.9} 
\begin{tabular}{lcccc}
\toprule
Model & Temperature & Top-$p$ & Top-$k$ & Max seq len \\
\midrule
Qwen3-VL-Thinking  & 0.6 & 0.95 & 20 & 51200 \\
Qwen3-VL-Instruct  & 1.0 & 1.0 & 0  & 51200 \\
DeepEyes   & 1.0 & 1.0 & 0  & 51200 \\
\bottomrule
\end{tabular}
\end{table}

\newpage

\section{Pseudocode}
\label{appendix: Pseudocode}
In this section, we provide pseudocode of the proposed RTWI under the online and offline settings.

\begin{algorithm*}[ht]
\caption{Reliable Online TWI}
\label{alg: online}
\begin{algorithmic}[1]
\STATE \textbf{Inputs:} Multimodal question $\{Q,I\}$, budget $B$, filtering ratio $\alpha$, threshold $\beta$, $i\leftarrow 0$.
\STATE \textbf{Offline Warmup:} 
\STATE Generate warmup traces $T_{\text{sel}}$ for $\{Q,I\}$ , initialize $(\eta^{m},\eta^{r})$ using Eq.~\ref{eq: dual-level thresholds} with $\alpha$, initialize $T_{\text{rel}}$ using Eq.~\ref{eq: dual-stage filtering}.
\STATE Compute voting values $V(a) = \sum_{t \in T_{\text{rel}}} C_t \cdot \mathbb{I}[A_t = a]$ for all answer $a$ and majority answer $\hat{a} = \arg\max_a V(a)$.
\STATE \textbf{Online Generation:}
\WHILE{$V(\hat{a}) / \sum_a V(a) < \beta$ \textbf{and} $|T_{\text{sel}}| + i < B$}
    \STATE \textbf{while} generating trace $t_i$ \textbf{do} 
    \STATE \quad Generate token $j$ and update reliability $w(t_i^s)$ for $s \in \{m,r\}$.
    \STATE \quad \textbf{If} $\ \exists \ s, \ w(t_i^s) < \eta^s$: stop generating $t_i$; \textbf{else}: add $j$ to $t_i$.
    \STATE \textbf{end while}
    \STATE Update $i\leftarrow i+1$. \textbf{If} $t_i$ is completed: update $T_{\text{rel}} \leftarrow T_{\text{rel}} \cup \{t_i\}$, $V(a)$ and $\hat{a}$.
\ENDWHILE
\STATE \textbf{return} Final answer $\hat{A}$.
\end{algorithmic}
\end{algorithm*}

\begin{algorithm*}[ht]
\caption{Reliable Offline TWI}
\label{alg: offline}
\begin{algorithmic}[1]
\STATE \textbf{Inputs:} Multimodal question $\{Q,I\}$, number of traces $B$, filtering ratio $\alpha$.
\STATE Initialize trace set $T \leftarrow \emptyset$.
\FOR{$i = 1$ \textbf{to} $B$}
    \STATE Generate a complete thinking trace $t_i$ for $\{Q,I\}$.
    \STATE $T \leftarrow T \cup \{t_i\}$.
\ENDFOR
\STATE Calculate thresholds $(\eta^{m},\eta^{r})$ using Eq.~\ref{eq: dual-level thresholds} with $T$ and $\alpha$, and then construct reliable trace set $T_{\text{rel}}$ using Eq.~\ref{eq: dual-stage filtering}.
\STATE Determine the final answer $\hat{A}$ using Eq.~\ref{eq: reliable_voting} with $T_{\text{rel}}$.
\STATE \textbf{return} Final answer $\hat{A}$.
\end{algorithmic}
\end{algorithm*}

\newpage

\section{More Experimental Results}
\label{appendix: more_experimental_results}
In this section, we present more experimental results of the proposed RTWI. Unless otherwise specified, all experiments are conducted using the Qwen3-VL-8B-Thinking model.

\subsection{More Experiments on Various TWI Models}
In the manuscript, we have carried out experiments using Qwen3-VL-8B-Thinking and Qwen3-VL-8B-Instruct.
To further verify the effectiveness of RTWI, we provide more experimental results under online setting using the widely-used TWI model DeepEyes-7B~\cite{deepeyes}.
From the results in Table~\ref{tab: real_world_deepeyes}, RTWI outperforms all baselines across various datasets in terms of reasoning accuracy and efficiency.

\begin{table*}[ht]
\centering
\caption{Results on real-world high-resolution benchmarks under online setting using DeepEyes-7B~\cite{deepeyes} as backbone. Base$^{*}$ indicates our reproduced performance of DeepEyes-7B.}
\label{tab: real_world_deepeyes}
\tablestyle{8pt}{0.9} 
\begin{tabular}{lcccccccccccc}
\toprule
 & \multicolumn{4}{c}{Vstar Bench} & \multicolumn{4}{c}{HR-Bench 4K} & \multicolumn{4}{c}{HR-Bench 8K} \\
 & \multicolumn{2}{c}{Attr} & \multicolumn{2}{c}{Spatial} & \multicolumn{2}{c}{FSP} & \multicolumn{2}{c}{FCP} & \multicolumn{2}{c}{FSP} & \multicolumn{2}{c}{FCP} \\
 \cmidrule(lr){2-3} \cmidrule(lr){4-5} \cmidrule(lr){6-7} \cmidrule(lr){8-9} \cmidrule(lr){10-11} \cmidrule(lr){12-13}
 Method & ACC & TSR & ACC & TSR & ACC & TSR & ACC & TSR & ACC & TSR & ACC & TSR \\
\midrule
GPT-4o & 72.2 & - & 60.5 & - & 66.8 & - & 63.3 & - & 60.8 & - & 58.5 & - \\
Thyme & 83.5 & - & 80.3 & - & 91.0 & - & 63.0 & - & 86.5 & - & 57.5 & - \\
\midrule
\addlinespace[2pt]
\rowcolor{gray!10} \multicolumn{13}{c}{\textit{DeepEyes-7B}} \\
\addlinespace[2pt]
Base$^{*}$ & 86.1 & - & 80.3 & - & 87.8 & - & 56.5 & - & 81.3 & - & 55.3 & - \\
SC & 86.1 & - & 80.3 & - & 89.5 & - & 58.0 & - & \textbf{86.0} & - & 57.3 & - \\
ASC & 85.2 & 60.0 & 80.3 & 58.1 & 89.5 & 58.9 & 58.0 & 48.9 & \textbf{86.0} & 58.1 & 57.3 & 47.4 \\
ESC & 86.1 & 55.0 & 80.3 & 48.2 & 89.5 & 55.8 & 58.0 & 31.3 & \textbf{86.0} & 47.2 & 57.3 & 24.7 \\
CISC & \textbf{87.0} & 55.4 & 80.3 & 51.8 & 89.8 & 56.4 & 56.8 & 37.1 & 85.8 & 50.6 & 56.8 & 34.0 \\
Deepconf & 86.1 & 58.2 & 79.0 & 56.8 & 89.5 & 53.9 & 58.0 & 48.0 & \textbf{86.0} & 53.7 & 58.3 & 43.9 \\
Self-Uncer. & 85.2 & 62.5 & 80.3 & 56.8 & 89.5 & 61.9 & 58.0 & 40.2 & 85.3 & 57.5 & 56.5 & 36.5 \\
\rowcolor{pink!30}Ours & \textbf{87.0} & \textbf{67.5} & \textbf{81.6} & \textbf{66.9} & \textbf{90.0} & \textbf{64.5} & \textbf{60.0} & \textbf{53.4} & \textbf{86.0} & \textbf{63.5} & \textbf{59.0} & \textbf{50.1} \\
\bottomrule
\end{tabular}
\end{table*}

\subsection{More Experiments on VisualProbe Benchmark}
In the manuscript, we have conducted experiments on the TWI-oriented TreeBench.
Here, we carry out additional experiments on the VisualProbe~\cite{minio3} benchmark.
As shown in Table~\ref{tab: online_visual_probe}, RTWI achieves higher reasoning accuracy and efficiency than most baselines across different difficulty levels.

\begin{table*}[htbp]
\centering
\caption{Results on VisualProbe Benchmark.}
\label{tab: online_visual_probe}
\tablestyle{10pt}{0.9} 
\begin{tabular}{lcccccc}
\toprule
 & \multicolumn{2}{c}{Easy} & \multicolumn{2}{c}{Medium} & \multicolumn{2}{c}{Hard} \\
 \cmidrule(lr){2-3} \cmidrule(lr){4-5} \cmidrule(lr){6-7}
 Method & ACC & TSR & ACC & TSR & ACC & TSR \\
\midrule
GPT-4o & 47.5 & - & 15.4 & - & 11.2 & - \\
DeepEyes & 60.1 & - & 29.8 & - & 35.1 & - \\
\midrule
\addlinespace[2pt]
\rowcolor{gray!10} \multicolumn{7}{c}{\textit{Qwen3-VL Thinking}} \\
\addlinespace[2pt]
Base & 47.6 & - & 25.8 & - & 28.3 & - \\
SC & 56.8 & - & 29.6 & - & 30.2 & - \\
ASC & 56.8 & 37.2 & 29.6 & 32.6 & 30.2 & 29.7 \\
ESC & 56.8 & 11.4 & 29.6 & 10.9 & 30.2 & 8.6 \\
CISC & 52.5 & 22.7 & 27.7 & 20.4 & 27.4 & 15.0 \\
Deepconf & 56.8 & 42.9 & 29.6 & 41.0 & 29.2 & 40.3 \\
Self-Cer. & 54.0 & 23.9 & 27.3 & 21.4 & 27.4 & 14.5 \\
\rowcolor{pink!30}Ours & \textbf{60.4} & \textbf{44.3} & \textbf{34.1} & \textbf{41.9} & \textbf{30.2} & \textbf{44.6} \\
\midrule
\addlinespace[2pt]
\rowcolor{gray!10} \multicolumn{7}{c}{\textit{Qwen3-VL Instruct}} \\
\addlinespace[2pt]
Base      & 56.0 & - & 35.4 & - & 34.9 & - \\
SC        & 64.0 & - & 40.2 & - & 42.5 & - \\
ASC       & 64.0 & \textbf{20.7} & 40.2 & \textbf{21.8} & 42.5 & \textbf{22.5} \\
ESC       & 64.0 & 4.7 & 40.2 & 8.8 & 42.5 & 6.7 \\
CISC      & 59.7 & 12.8 & 38.6 & 14.7 & 39.6 & 9.2 \\
Deepconf  & 61.9 & 14.6 & \textbf{40.6} & 16.7 & \textbf{43.4} & 14.7 \\
Self-Cer. & 59.0 & 13.9 & 39.4 & 14.5 & 37.7 & 12.0 \\
\rowcolor{pink!30}Ours & \textbf{65.4} & 16.2 & \textbf{40.6} & 18.5 & 42.5 & 16.2 \\ \hline
\bottomrule
\end{tabular}
\end{table*}

\subsection{Offline Evaluations}
\label{appendix: offline_evaluation}
In the manuscript, we have conducted experiments under online setting. 
Here, we further provide more implementation details and experimental results on various benchmarks under offline setting.
Specifically, Alg.~\ref{alg: offline} provides the details of the algorithm under offline setting.
From the results in Table~\ref{tab: real_world_offline}-\ref{table: multimodal_reasoning_offline}, one could have the following conclusions:
i) RTWI consistently outperforms all baselines across different benchmarks under offline setting, demonstrating its robustness against NT;
ii) comparing performances between the online and offline settings, the online implementation achieves similar results to its offline counterpart, indicating that the proposed warmup strategy could effectively estimate the reliability thresholds and thus facilitate dual-stage filtering and voting.

\begin{table}[ht]
\centering
\caption{Results on real-world high-resolution benchmarks under offline setting.}
\label{tab: real_world_offline}
\tablestyle{12pt}{0.9} 
\begin{tabular}{lcccccc}
\toprule
 & \multicolumn{2}{c}{Vstar Bench} & \multicolumn{2}{c}{HR-Bench 4K} & \multicolumn{2}{c}{HR-Bench 8K} \\
 \cmidrule(lr){2-3} \cmidrule(lr){4-5} \cmidrule(lr){6-7}
 Method & Attr & Spatial & FSP & FCP & FSP & FCP \\
\midrule
\rowcolor{gray!10} \multicolumn{7}{c}{\textit{Qwen-VL Thinking}} \\
\addlinespace[2pt]
Base      & 78.3 & 73.7 & 83.0 & 57.3 & 78.3 & 56.0 \\
SC        & 80.0 & 77.6 & 89.5 & 61.0 & 86.5 & 61.0 \\
CISC      & 84.3 & 69.7 & 87.8 & 59.8 & 88.5 & 56.0 \\
Deepconf  & 81.7 & 76.3 & 89.5 & 62.5 & 87.0 & 61.8 \\
Self-Cer. & 82.6 & 68.4 & 89.5 & 61.0 & 88.0 & 59.0 \\
\rowcolor{pink!30}Ours      & \textbf{86.1} & \textbf{82.9} & \textbf{91.5} & \textbf{65.5} & \textbf{89.0} & \textbf{63.0} \\
\midrule
\rowcolor{gray!10} \multicolumn{7}{c}{\textit{Qwen-VL Instruct}} \\
\addlinespace[2pt]
Base      &  90.4 & 86.8 & 93.8 & 72.0 & 90.8 & 71.3 \\
SC        & 92.2 & 89.5 & 95.3 & 77.0 & 93.0 & 74.8 \\
CISC      & 93.9 & 89.5 & 95.8 & 77.5 & 93.0 & 75.8 \\
Deepconf  & 93.9 & \textbf{90.8} & 96.0 & 76.5 & 93.0 & 75.8 \\
Self-Cer. & 93.9 & \textbf{90.8} & 96.0 & 76.5 & 92.5 & 75.5 \\
\rowcolor{pink!30}Ours      & \textbf{94.8} & \textbf{90.8} & \textbf{96.5} & \textbf{78.8} & \textbf{95.0} & \textbf{76.0} \\
\bottomrule
\end{tabular}
\end{table}

\begin{table*}[ht]
\centering
\begin{minipage}[t]{0.5\textwidth} 
    \vspace{0pt}
    \tablestyle{10pt}{0.9}{ 
        \centering
        \caption{Results on TWI-oriented TreeBench under offline setting.} 
        \label{table: twi_oriented_offline}
        \begin{tabular}{lcccc}
\toprule
 & \multicolumn{2}{c}{Reasoning} & \multicolumn{2}{c}{Perception} \\
 \cmidrule(lr){2-3} \cmidrule(lr){4-5}
 Method & ACC & mIoU & ACC & mIoU \\
\midrule
\rowcolor{gray!10} \multicolumn{5}{c}{\textit{Qwen3-VL Thinking}} \\
\addlinespace[2pt]
Base      & 32.8 & - & 63.1 & - \\
SC        & 30.9 & 37.8 & 66.4 & 32.5 \\
CISC      & 25.8 & 37.8 & 66.4 & 32.5 \\
Deepconf  & 30.9 & 39.0 & 64.4 & 32.9 \\
Self-Cer. & 30.9 & 37.8 & 65.8 & 32.5 \\
\rowcolor{pink!30} Ours & \textbf{37.1} & \textbf{41.8} & \textbf{68.5} & \textbf{36.0} \\
\midrule
\rowcolor{gray!10} \multicolumn{5}{c}{\textit{Qwen3-VL Instruct}} \\
\addlinespace[2pt]
Base      &  34.4 & - & 64.4 & - \\
SC        & 34.8 & 43.4 & 68.4 & 38.6 \\
CISC      & 34.8 & 43.4 & 68.4 & 38.6 \\
Deepconf  & 35.6 & 43.6 & 69.8 & 38.9 \\
Self-Cer. & 35.6 & 43.4 & 69.8 & 38.6 \\
\rowcolor{pink!30} Ours & \textbf{35.9} & \textbf{46.8} & \textbf{71.1} & \textbf{42.0} \\
\bottomrule
\end{tabular}
    }
\end{minipage}
\hspace{5pt}
\begin{minipage}[t]{0.46\textwidth} 
    \vspace{0pt}
    \tablestyle{6pt}{0.93}{
        \centering
        \caption{Results on reasoning benchmarks under offline setting.}
        \label{table: multimodal_reasoning_offline}
        \begin{tabular}{lcc}
\toprule
Method & MathVision & LogicVista \\
\midrule
\rowcolor{gray!10} \multicolumn{3}{c}{\textit{Qwen3-VL Thinking}} \\
\addlinespace[2pt]
Base      & 18.7 & 44.3 \\
SC        & 19.8 & 46.4 \\
CISC      & 20.1 & 50.2 \\
Deepconf  & 19.7 & 50.8 \\
Self-Cer. & 18.0 & 49.1 \\
\rowcolor{pink!30} Ours & \textbf{23.4} & \textbf{59.1} \\
\midrule
\rowcolor{gray!10} \multicolumn{3}{c}{\textit{Qwen3-VL Instruct}} \\
\addlinespace[2pt]
Base      & 21.4 & 47.9 \\
SC        & 24.4 & 54.8 \\
CISC      & 22.7 & 55.3 \\
Deepconf  & 22.7 & 55.4 \\
Self-Cer. & 24.0 & 55.4 \\
\rowcolor{pink!30} Ours & \textbf{25.5} & \textbf{60.3} \\
\bottomrule
\end{tabular}
    }
\end{minipage}
\end{table*}

\subsection{Analytic Study on Tool Usage}
\begin{figure}[htbp]
    \centering
    \includegraphics[width=0.7\linewidth]{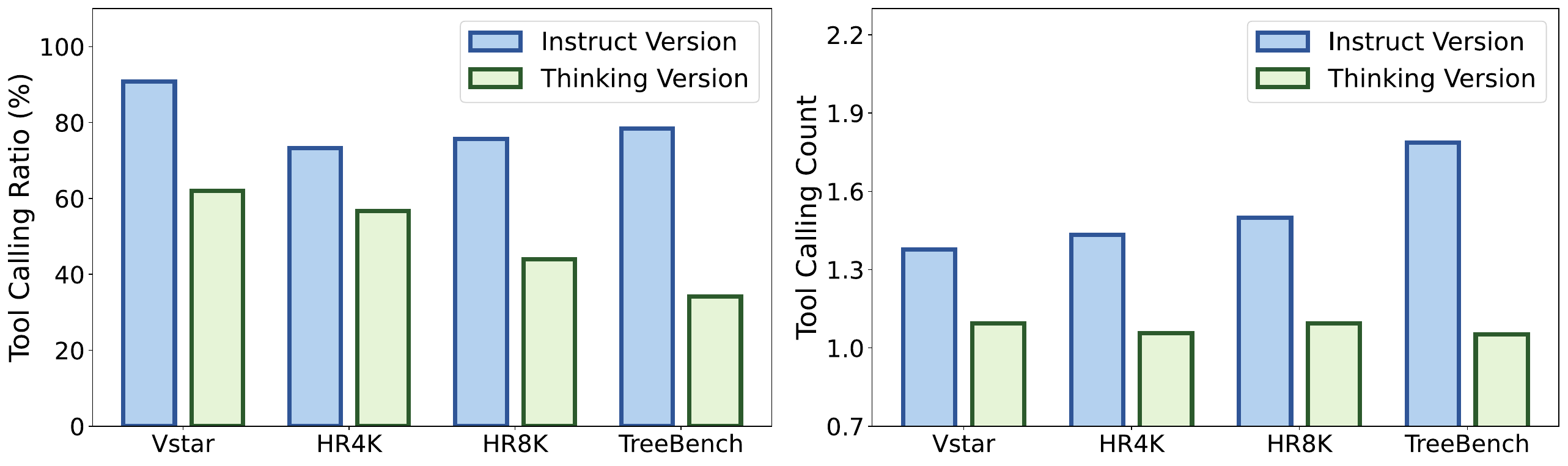}
    \caption{Analytic study on tool usage.}
    \label{fig: tool_usage}
\end{figure}

As mentioned in Section~\ref{sec: single-stage reliability}, not all generated traces involve tool usage.
Here, we conduct an additional analytical study to further examine tool usage with Qwen3-VL-8B-Thinking and Qwen3-VL-8B-Instruct as the TWI model.
As illustrated in Fig.~\ref{fig: tool_usage}, the Instruct variant exhibits stronger instruction-following ability than the Thinking variant, leading to more frequent tool usage with a higher number of invocations.
Notably, such a phenomenon indicates that the proposed RTWI could improve performance for both textual CoT and interleaved CoT with tool invocations.

\subsection{Analytic Study on CoT Length}
\label{appendix: trace_length}
\begin{figure}[ht]
    \centering
    \includegraphics[width=0.8\linewidth]{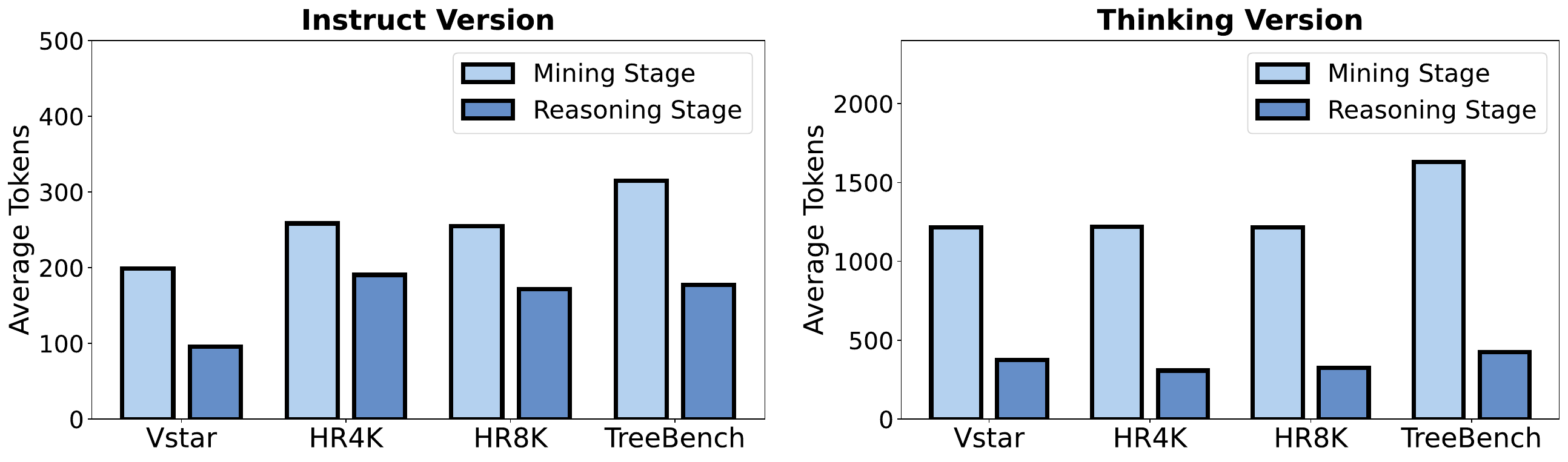}
    \caption{Analytic study on CoT length.}
    \label{fig: trace_length}
\end{figure}

As pointed out in Section~\ref{sec: reliability_estimation}, the lengths of textual CoT in the mining and reasoning stages are not necessarily consistent.
Here, we conduct additional analytical experiments to examine stage-wise CoT lengths. 
As shown in Fig.~\ref{fig: trace_length}, one could have the following observations:
i) the same TWI model generates textual CoTs of varying lengths across different datasets, \textit{e.g.}, on more challenging benchmarks such as TreeBench, the model tends to generate more tokens;
ii) the CoT length in the mining stage is generally longer than that in the reasoning stage, suggesting that TWI models allocate substantial computation to acquiring visual cues for facilitating the subsequent reasoning process;
iii) the Thinking variant generates longer textual CoTs than the Instruct variant.
Based on these observations, the proposed Top-$k$ selection strategy with a self-adaptively determined $k$ enables more stable reliability estimation across different datasets and TWI models.

\subsection{Analytic Study on Saving Tokens}
\begin{figure}[ht]
    \centering
    \includegraphics[width=0.95\linewidth]{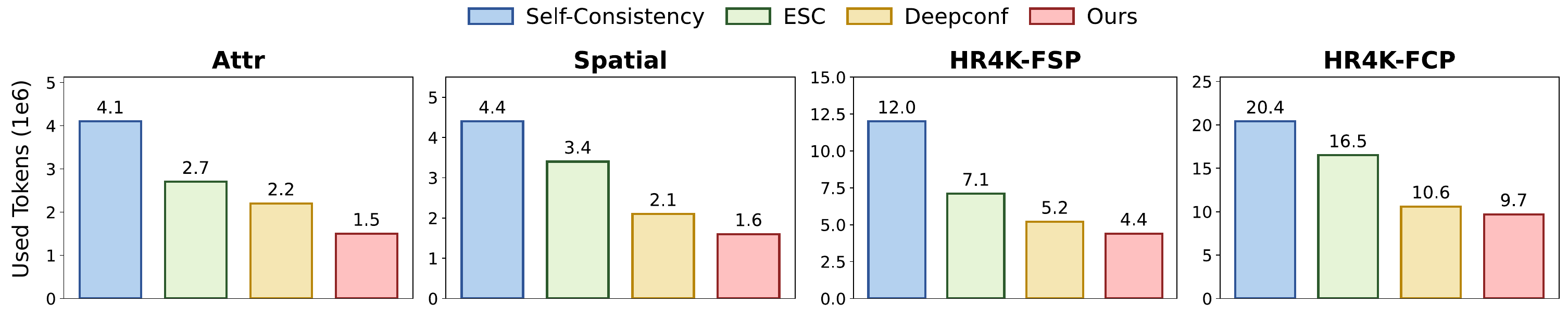}
    \caption{Generated tokens comparison across different benchmarks.}
    \label{fig: saving_toknes}
\end{figure}

Here, we visualize token reduction patterns across various benchmarks in Fig.~\ref{fig: saving_toknes}, illustrating how RTWI achieves substantial computational savings while preserving competitive accuracy.

\newpage

\section{More Ablation Studies}
In this section, we present more ablation studies about the parameter analysis.

\subsection{Ablation Study on Temperature}
\begin{figure}[ht]
    \centering
    \includegraphics[width=0.7\linewidth]{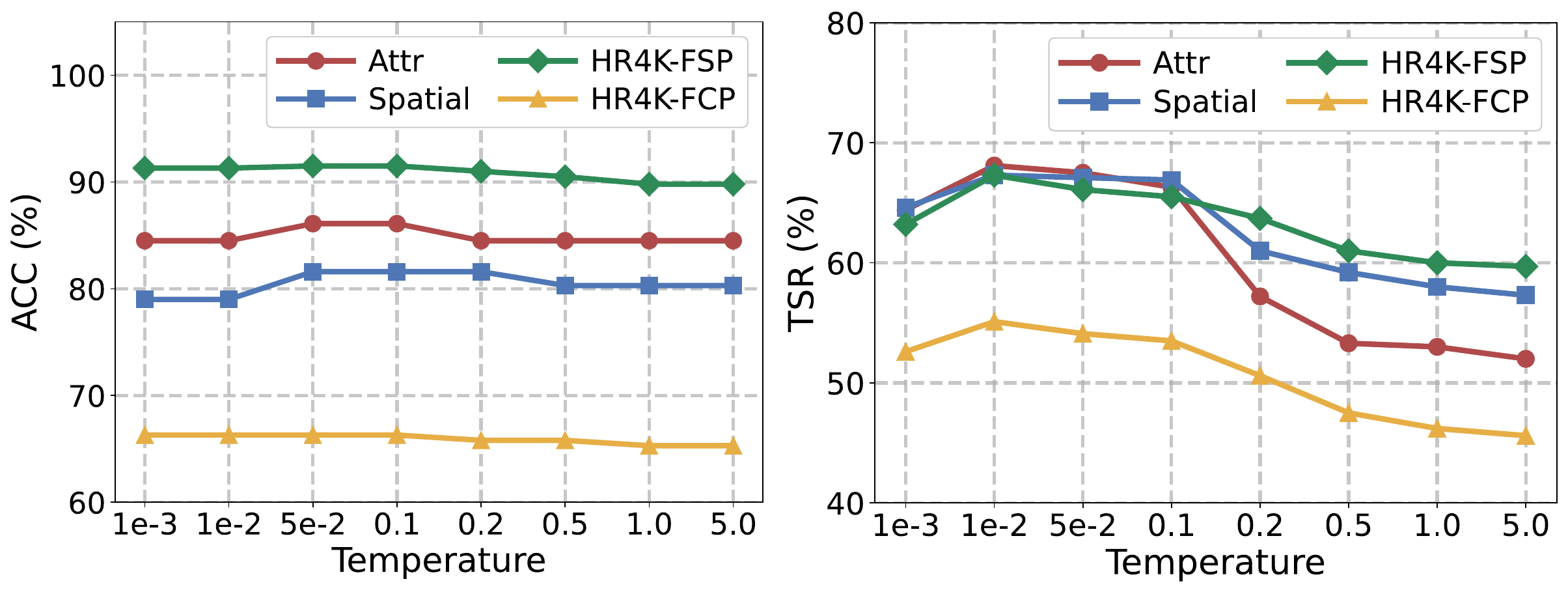}
    \caption{Ablation study on temperature $\tau$ in Eq.~\ref{eq: reliable_voting}.}
    \label{fig: ablation_tem}
\end{figure}

We carry out an additional ablation study on the temperature $\tau$ in Eq.~\ref{eq: reliable_voting}.
From the results in Fig.~\ref{fig: ablation_tem}, we observe that RTWI exhibits stable performance and token saving ratios when $\tau$ falls within the range of $[0.05, 0.1]$.

\subsection{Ablation on Warmup Traces}
We conduct an additional ablation study on the number of warm-up traces $|T_{\text{sel}}|$.
As shown in Table.~\ref{tab: ablation_warmup}, we observe that increasing $|T_{\text{sel}}|$ generally makes reasoning accuracy under the online setting closer to that under the offline setting, while reducing token saving ratios as RTWI could only employ early stopping on the remaining $B - |T_{\text{sel}}|$ traces.

\begin{table*}[ht]
\centering
\caption{\textbf{Ablation Study on the Number of Warmup Traces $|T_{\text{sel}}|$.}}
\label{tab: ablation_warmup}
\tablestyle{8pt}{1.0} 
\begin{tabular}{llcccccccc}
\toprule
 & & \multicolumn{4}{c}{Vstar Bench} & \multicolumn{4}{c}{HR-Bench 4K} \\
 & & \multicolumn{2}{c}{Attr} & \multicolumn{2}{c}{Spatial} & \multicolumn{2}{c}{FSP} & \multicolumn{2}{c}{FCP} \\
 \cmidrule(lr){3-4} \cmidrule(lr){5-6} \cmidrule(lr){7-8} \cmidrule(lr){9-10}
 $|T_{\text{sel}}|$ & Method & ACC & TSR & ACC & TSR & ACC & TSR & ACC & TSR \\
\midrule
\multirow{2}{*}{4.0} & Deepconf & 80.0 & 64.0 & 77.6 & 67.5 & 89.0 & 63.6 & 60.3 & 64.9 \\
 & Ours & 86.1 & 74.6 & 80.3 & 72.6 & 90.0 & 75.3 & 64.0 & 66.3 \\
\midrule
\multirow{2}{*}{8.0} & Deepconf & 81.7 & 47.6 & 76.3 & 52.8 & 90.5 & 57.3 & 62.5 & 48.9 \\
 & Ours & 85.2 & 61.1 & 81.6 & 61.2 & 91.3 & 61.9 & 65.8 & 50.5 \\
\midrule
\multirow{2}{*}{12.0} & Deepconf & 81.7 & 44.2 & 76.3 & 46.4 & 90.5 & 48.6 & 62.0 & 30.5 \\
 & Ours & 86.1 & 48.2 & 82.9 & 51.4 & 91.5 & 53.8 & 66.0 & 31.7 \\
\midrule
\multirow{2}{*}{16.0} & Deepconf & 81.7 & 36.7 & 77.6 & 37.5 & 91.5 & 27.9 & 62.5 & 26.2 \\
 & Ours & 86.1 & 38.5 & 81.6 & 41.1 & 91.5 & 30.4 & 66.0 & 30.1 \\
\bottomrule
\end{tabular}
\end{table*}

\newpage

\section{Evaluation Prompts}
To ensure reproducibility and facilitate future research, we provide the prompts used to evaluate RTWI across all benchmarks.
Note that the prompts are adapted from the official prompts of Qwen3-VL~\cite{Qwen3-VL}.

\begin{minipage}{0.99\columnwidth}
\begin{tcolorbox} 
    \raggedright
    \small
    \textbf{Prompt for Multiple-Choice Benchmarks}
    \begin{itemize}[leftmargin=4.5mm]
    \setlength{\itemsep}{2pt}
    
    \item Your role is that of a research assistant specializing in visual information. Answer questions about images by looking at them closely and providing detailed analysis. Please follow this structured thinking process and show your work.

    \item Start an iterative loop for each question:
    \begin{itemize}[leftmargin=4.5mm]
        \setlength{\itemsep}{2pt}
        \item \textbf{First, look closely:} Begin with a detailed description of the image, paying attention to the user's question. List what you can tell just by looking, and what you'll need to look up.
        \item \textbf{Next, find information:} Use a tool to research the things you need to find out.
        \item \textbf{Then, review the findings:} Carefully analyze what the tool tells you and decide on your next action.
    \end{itemize}

    \item Continue this loop until your research is complete.

    \item To finish, put your final answer within \textbackslash boxed\{\}. Answer with the option's letter from the given choices directly, e.g. \textbackslash boxed\{A\}, \textbackslash boxed\{B\}, \textbackslash boxed\{C\}, \textbackslash boxed\{D\}, etc.
    \end{itemize}
\end{tcolorbox}
\end{minipage}

\begin{minipage}{0.99\columnwidth}
\begin{tcolorbox} 
    \raggedright
    \small
    \textbf{Prompt for Open-Ended Benchmarks}
    \begin{itemize}[leftmargin=4.5mm]
    \setlength{\itemsep}{2pt}
    
    \item Your role is that of a research assistant specializing in visual information. Answer questions about images by looking at them closely and providing detailed analysis. Please follow this structured thinking process and show your work.

    \item Start an iterative loop for each question:
    \begin{itemize}[leftmargin=4.5mm]
        \setlength{\itemsep}{2pt}
        \item \textbf{First, look closely:} Begin with a detailed description of the image, paying attention to the user's question. List what you can tell just by looking, and what you'll need to look up.
        \item \textbf{Next, find information:} Use a tool to research the things you need to find out.
        \item \textbf{Then, review the findings:} Carefully analyze what the tool tells you and decide on your next action.
    \end{itemize}

    \item Continue this loop until your research is complete.

    \item To finish, You MUST PUT your FINAL ANSWER within \textbackslash boxed\{\}, and make sure it contains only the answer itself without extra words or symbols.
    \end{itemize}
\end{tcolorbox}
\end{minipage}

\newpage

\section{Case Study}
\label{appendix: case_study}
\begin{figure}[ht]
    \centering
    \includegraphics[width=0.6\linewidth]{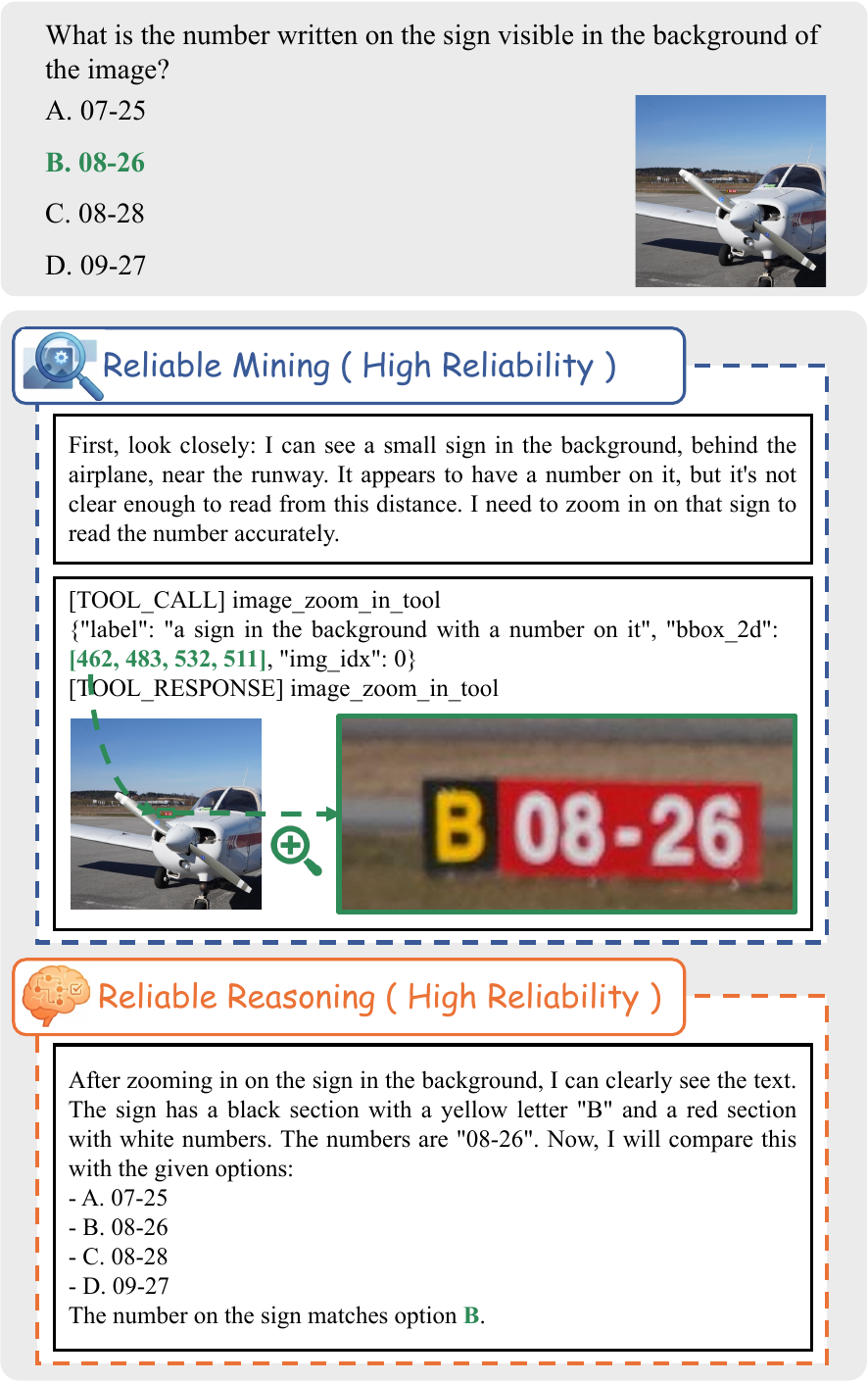}
    \caption{Example of reliable trace: visual perception and OCR. The TWI model correctly perceives the sign and performs OCR to recognize the numbers displayed on it.}
    \label{fig: case_reliable_1}
\end{figure}

\begin{figure}[ht]
    \centering
    \includegraphics[width=0.6\linewidth]{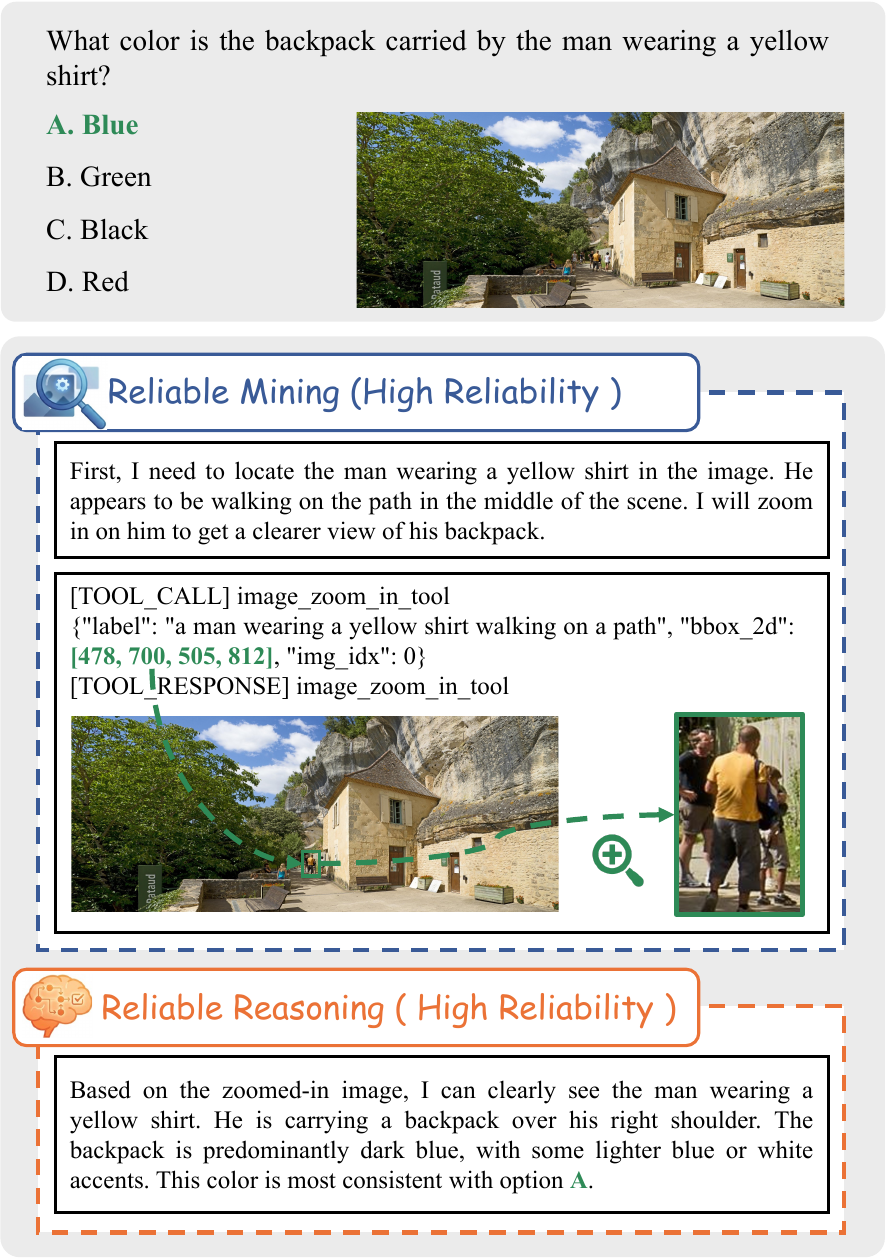}
    \caption{Example of reliable trace: visual perception and color identification. The TWI model identifies the man wearing a yellow shirt and correctly recognizes the color of his backpack.}
    \label{fig: case_reliable_2}
\end{figure}

\begin{figure}[ht]
    \centering
    \includegraphics[width=0.6\linewidth]{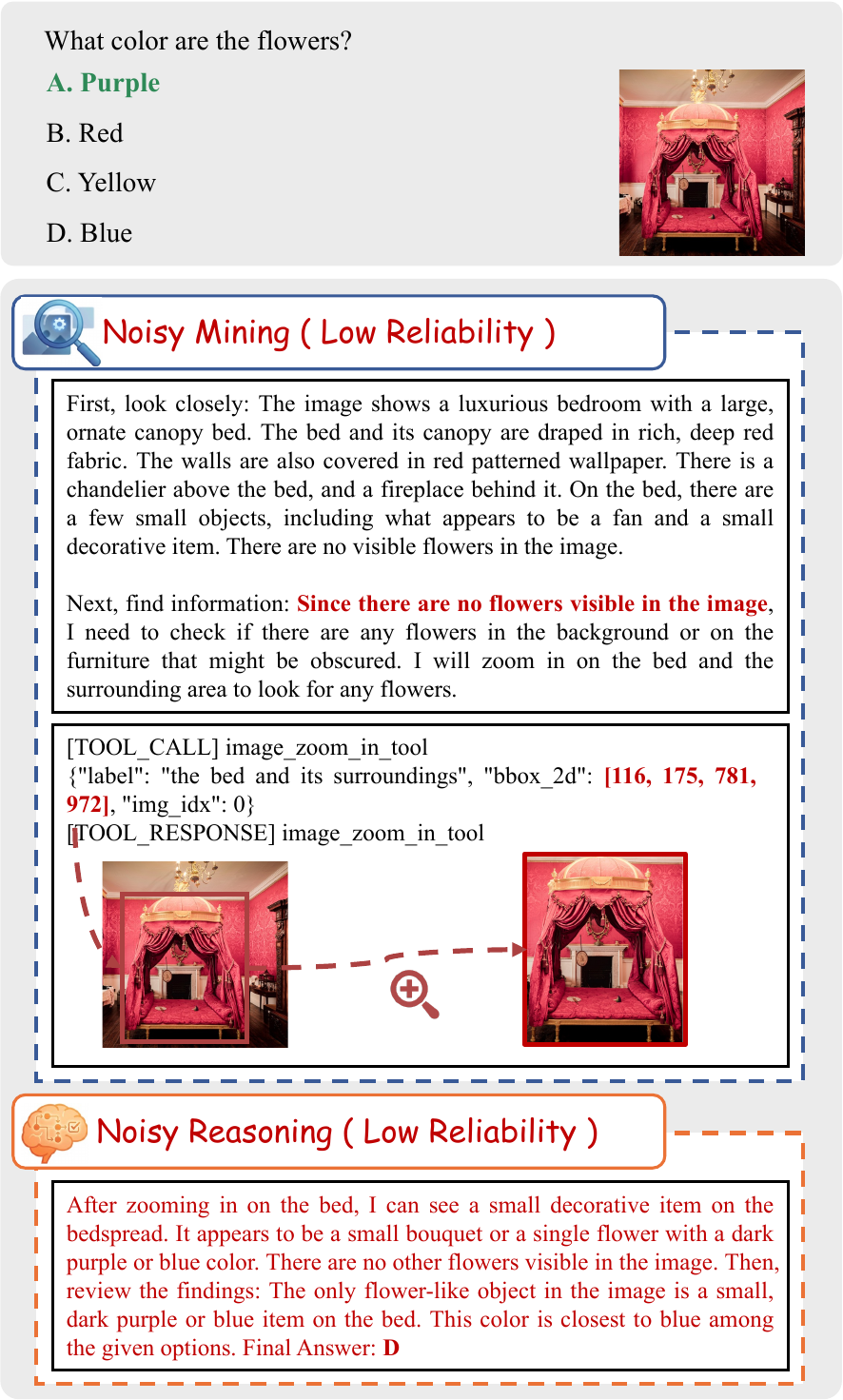}
    \caption{Example of noisy mining: failed visual perception. The TWI model locates the coarse-grained region but fails to derive the fine-grained flowers required for correct reasoning.}
    \label{fig: case_noisy_mining_1}
\end{figure}

\begin{figure}[ht]
    \centering
    \includegraphics[width=0.5\linewidth]{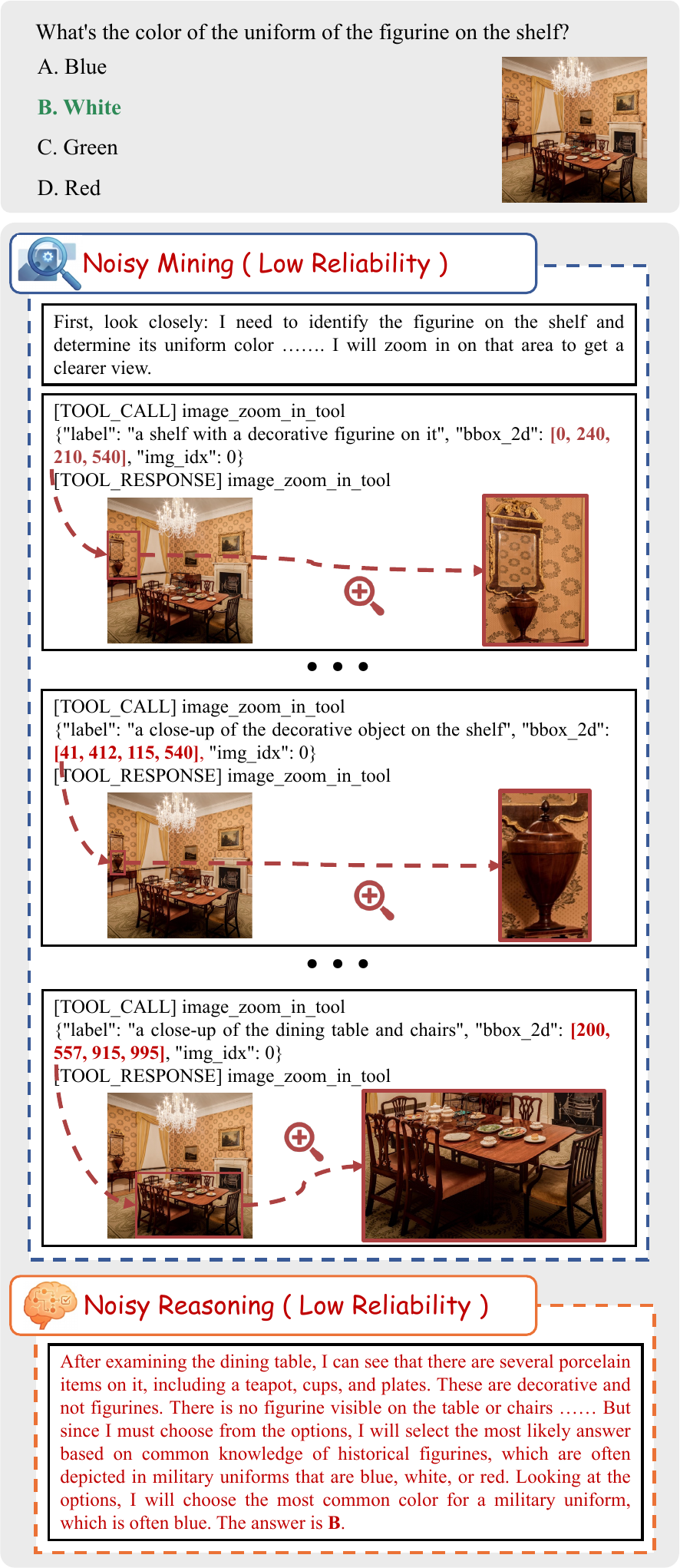}
    \caption{Example of noisy mining: failed visual search. The TWI model repeatedly searches the image but fails to locate the figurine in the thinking stage.}
    \label{fig: case_noisy_mining_2}
\end{figure}

\begin{figure}[ht]
    \centering
    \includegraphics[width=0.6\linewidth]{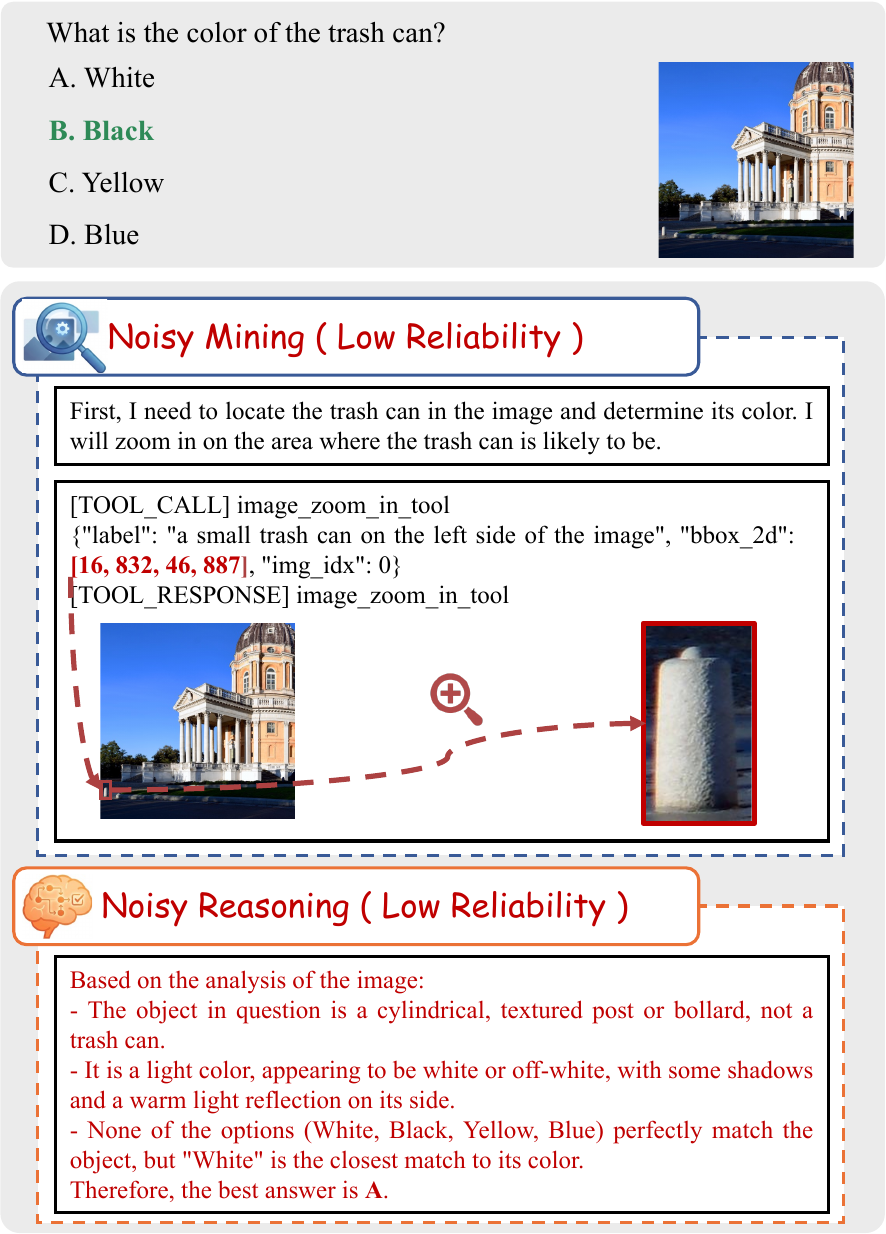}
    \caption{Example of noisy mining: false positive visual perception. The TWI model mistakenly identifies a stone pillar as a trash can in the mining stage, then recognizes the visual cue is incorrect yet still derives the wrong answer during the reasoning stage.}
    \label{fig: case_noisy_mining_3}
\end{figure}

\begin{figure}[ht]
    \centering
    \includegraphics[width=0.55\linewidth]{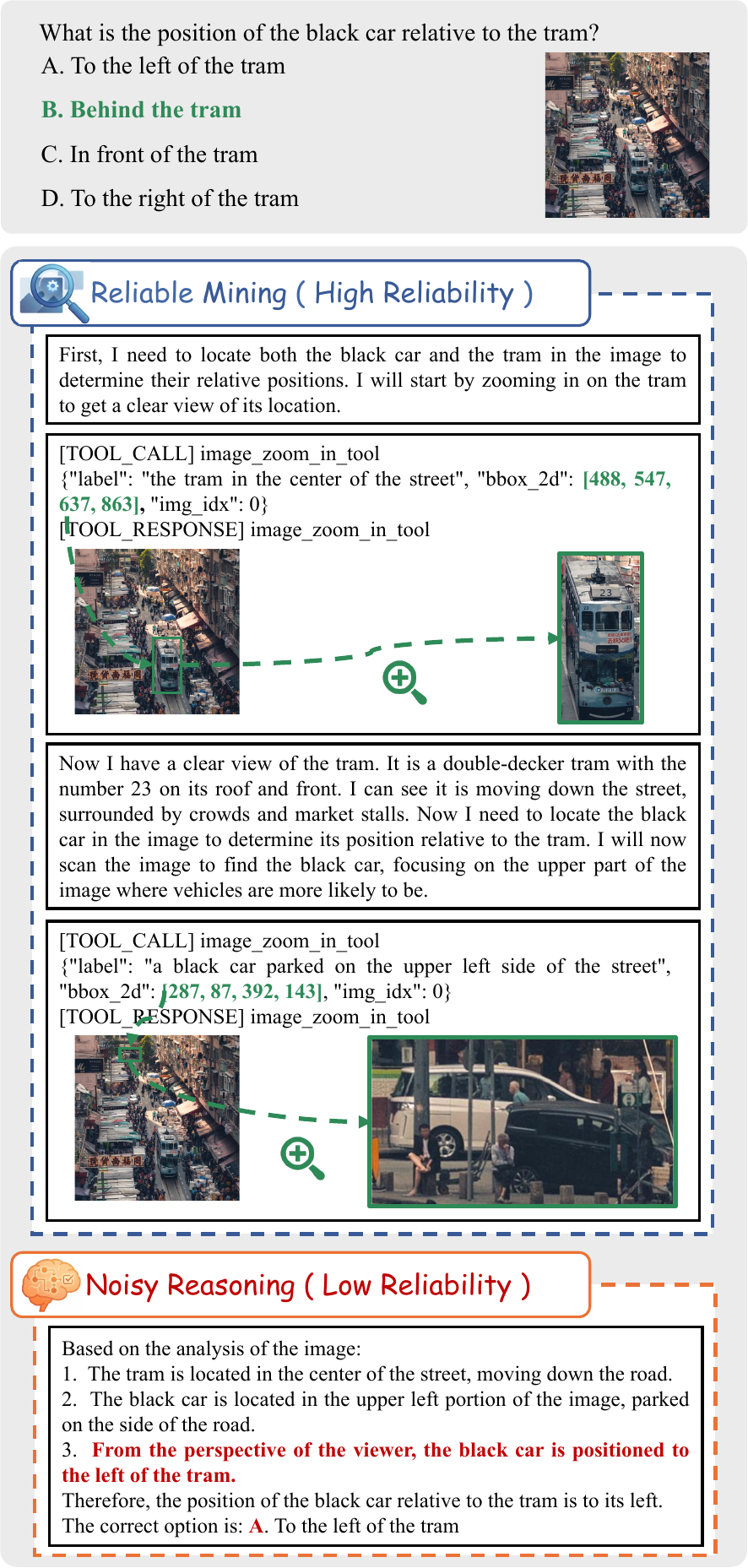}
    \caption{Example of noisy reasoning: limited understanding of perspective projection. Although the TWI model correctly locates both the tram and the black car, it fails to account for the visual perspective and spatial mapping between them in the reasoning stage.}
    \label{fig: case_noisy_reasoning_1}
\end{figure}

\begin{figure}[ht]
    \centering
    \includegraphics[width=0.6\linewidth]{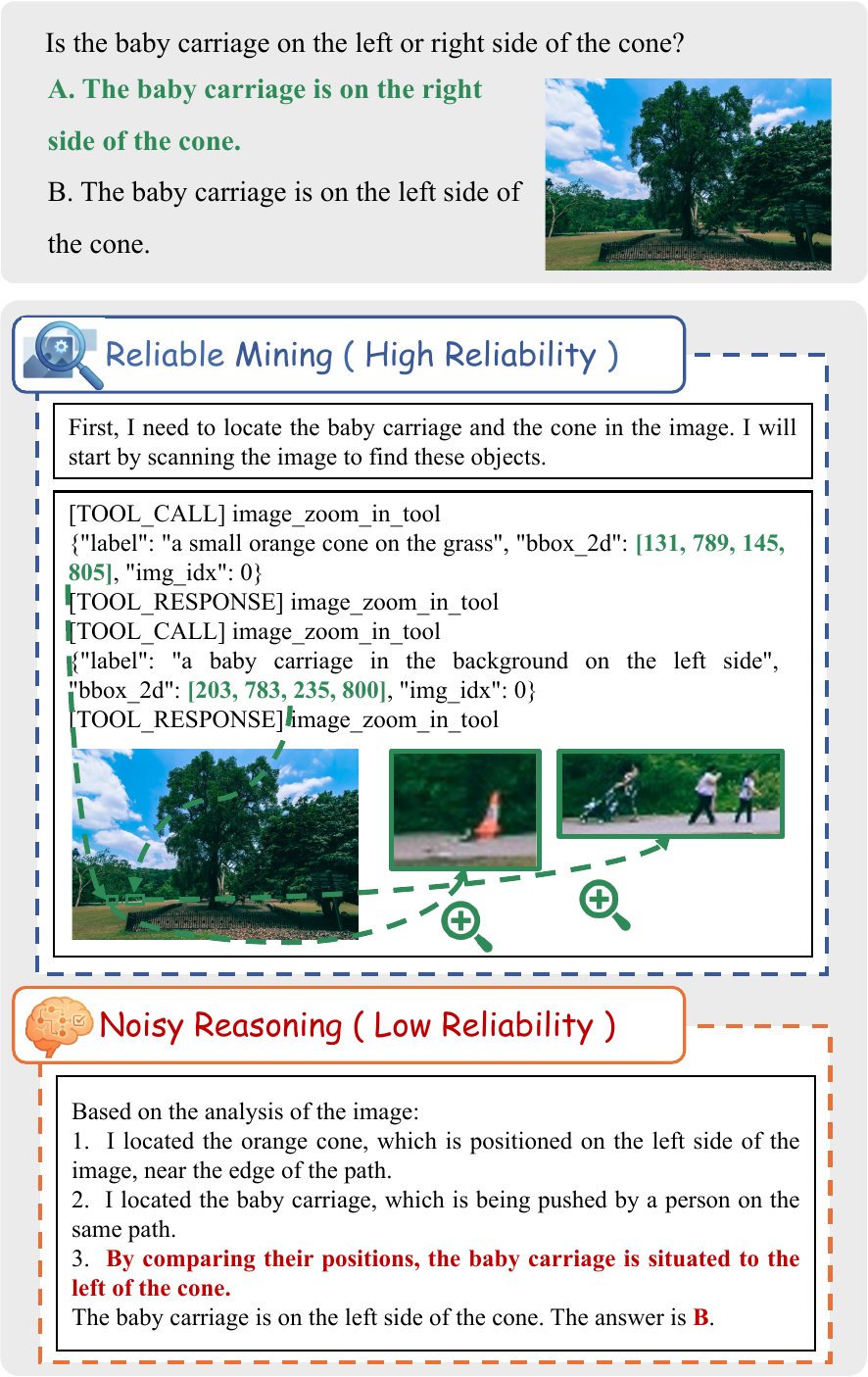}
    \caption{Example of noisy reasoning: limited spatial understanding. Although the TWI model successfully locates both the baby carriage and the cone, it misinterprets the spatial relationship in the reasoning stage.}
    \label{fig: case_noisy_reasoning_2}
\end{figure}

\begin{figure}[ht]
    \centering
    \includegraphics[width=0.6\linewidth]{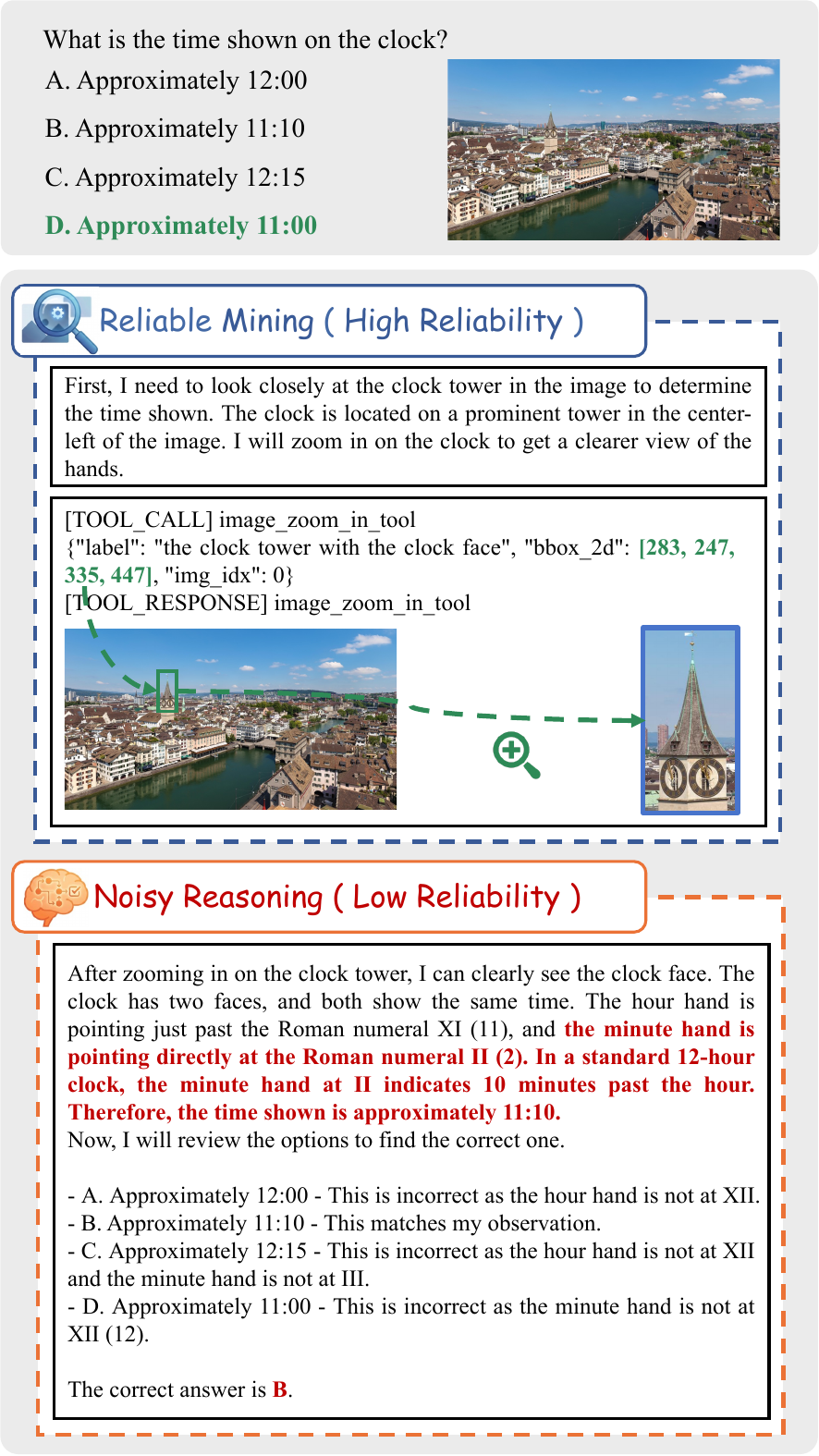}
    \caption{Example of noisy reasoning: symbolic misinterpretation. Although the TWI model successfully localizes the clock and zooms in on the clock face, it misinterprets the clock symbols during the reasoning stage.}
    \label{fig: case_noisy_reasoning_3}
\end{figure}

In this section, we provide some examples of successful identification of NT, as well as cases of failure.
As shown in Fig.~\ref{fig: case_reliable_1}–\ref{fig: case_noisy_reasoning_3}, RTWI could effectively identify reliable and noisy thinking processes by estimating reliability for both the visual cues in the mining stage and the textual CoT in the reasoning stage.

\end{document}